\documentclass[10pt]{article}
\usepackage[preprint]{tmlr}

\usepackage{amsmath,amsfonts,bm}

\def\eqref#1{equation~\ref{#1}}

\def\1{\bm{1}}

\DeclareMathAlphabet{\mathsfit}{\encodingdefault}{\sfdefault}{m}{sl}
\SetMathAlphabet{\mathsfit}{bold}{\encodingdefault}{\sfdefault}{bx}{n}

\usepackage{amsmath,amssymb,amsfonts}
\usepackage{wrapfig}
\makeatletter
\newcommand\wrapfill{\par
  \ifx\parshape\WF@fudgeparshape
    \nobreak\vskip-\baselineskip
    \vskip\c@WF@wrappedlines\baselineskip
    \allowbreak\WFclear
  \fi}
\makeatother
\usepackage{enumitem}
\usepackage{graphicx}
\usepackage{textcomp}
\usepackage{algorithm}
\usepackage{algpseudocode}
\usepackage{xcolor}
\usepackage{hyperref}
\hypersetup{hidelinks}
\usepackage{url}
\usepackage{tabularx}
\usepackage{booktabs}  
\usepackage{array}     
\usepackage{booktabs}
\usepackage{placeins} 
\usepackage{needspace} 

\usepackage{dsfont}
\newcommand{\ind}{\mathds{1}}

\usepackage{etoolbox}
\AtBeginEnvironment{table}{\small\centering}
\AtBeginEnvironment{wraptable}{\footnotesize\centering}

\title{Graph-Based Dynamic Expert Selection for Personalized Federated Learning}

\author{\name Brianna Mueller \email brianna-mueller@uiowa.edu \\
      \addr Department of Business Analytics\\
      University of Iowa\\
      Iowa City, USA
      \AND
      \name W. Nick Street \email nick-street@uiowa.edu \\
      \addr Department of Business Analytics\\
      University of Iowa\\
      Iowa City, USA}

\def\month{MM}
\def\year{YYYY}
\def\openreview{\url{https://openreview.net/forum?id=XXXX}}

\begin{document}

\maketitle

\begin{abstract}

Personalized federated learning (pFL) aims to mitigate negative transfer that arises when a global model is shared across clients with heterogeneous data. Yet most pFL methods consolidate cross-client knowledge before it reaches a target client, so knowledge is transferred from all source clients regardless of their relevance. Beyond this coarse treatment of inter-client heterogeneity, prior work has largely overlooked intra-client heterogeneity, focusing on client-level personalization even though the relevance of source clients can vary across a target client's samples. The few strategies that provide finer-grained personalization through selective knowledge transfer or instance-level adaptation often assume a common model architecture, precluding the use of the models best suited to each client. To address these limitations, we introduce FedDES (Federated Dynamic Expert Selection), an instance-level pFL approach that treats clients' heterogeneous models as a pool of experts. In FedDES, client-specific graph neural networks learn joint representations of sample--expert pairs, which are used to route each input to a subset of experts, forming a weighted, input-dependent ensemble. Across benchmark datasets with simulated heterogeneity and real-world biomedical datasets with naturally occurring heterogeneity, FedDES consistently outperforms model-heterogeneous pFL baselines while most effectively mitigating negative transfer.

\end{abstract}

\section{Introduction}

Federated Learning (FL) is a distributed machine learning approach that enables multiple institutions or edge devices to collaboratively train models without sharing raw data. By keeping sensitive data localized, FL addresses privacy and regulatory constraints in applications such as healthcare diagnostics, financial fraud detection, and autonomous systems. Standard federated learning approaches such as Federated Averaging (FedAvg) \citep{mcmahan2017communication} train a single global model by iteratively aggregating local gradient updates from participating clients. However, the assumption that a single shared model can adequately serve all clients often fails in practice when clients' data are non-independent and identically distributed (non-IID), also referred to as \textbf{statistical heterogeneity}. Under such heterogeneity, forcing diverse clients to converge toward a shared representation can degrade performance for individual clients, a phenomenon known as \textbf{negative transfer}.

In response to this challenge, personalized federated learning (pFL) has emerged as a promising alternative that aims to provide each client with a model tailored to its local data distribution while still benefiting from collaborative learning. At its core, personalization is a problem of effective knowledge transfer: leveraging useful cross-client knowledge for local learning while avoiding negative transfer from distributional mismatch. Yet most existing approaches can undermine this goal by consolidating cross-client knowledge into a single global representation before it reaches a target client, so that knowledge is transferred from all source clients regardless of their relevance. These methods balance local and shared global knowledge through mechanisms such as local adaptation \citep{fallah2020personalized}, regularization toward a global reference \citep{t2020personalized,li2021ditto}, or local and global model interpolation \citep{deng2020adaptive}. This coarse-grained knowledge transfer can be particularly limiting or harmful in cross-silo federations, where statistical heterogeneity is often more pronounced because clients are organizations whose data reflect distinct populations and data collection practices \citep{huang2022crosssilo,zhang2025fairness,pye2021personalized}. Some methods transfer knowledge more selectively. Client clustering approaches limit collaboration to a group of similar clients \citep{sattler2020clustered}, while personalized aggregation methods learn client-specific weights that determine the influence of each source client based on estimated relevance \citep{zhang2020personalized,huang2021personalized}.

A second dimension of personalization granularity is the level at which personalization takes place. Across existing pFL methods, this has predominantly been the client level, where collaborative learning yields a single personalized predictor per client that is applied uniformly to all of its samples. Yet, just as the relevance of source clients can vary from one target client to another, it can also vary across a target client's samples \citep{feng2023fedins,panchal2023flow}. For example, a patient group that is underrepresented at one hospital may be well represented at another, making that hospital's knowledge especially valuable for those patients. For the patients it treats most often, the first hospital's own model may be the more reliable. A client-level approach may still benefit the hospital on average, producing a predictor that generalizes better than one trained independently. However, because overall performance is dominated by the most commonly observed patient profiles, the resulting predictor can be suboptimal for the underrepresented group, for whom a different combination of source clients may be more relevant.

A smaller, more recent body of work has explored instance-level personalization, primarily by adapting the balance between local and global knowledge for each sample \citep{zhang2023fedcp,panchal2023flow,yi2026pfedmoe}. However, we identify two ways in which this approach to per-sample adaptation can fall short. First, it implicitly assumes that, for any given sample, the most relevant knowledge in the federation is the client's local knowledge, the shared global knowledge, or some combination of the two. This may hold when clients differ mainly in how they sample a shared underlying distribution. When they instead form distinct subpopulations, as with the domain shift common in cross-silo federations \citep{zhang2025fairness}, a sample atypical of its client may resemble the data of only a small subset of other clients. In such cases, per-sample adaptation cannot restore what aggregation with the remaining clients has diluted. 

Second, for the samples that are best served by the client's own knowledge, the local component does not purely represent that knowledge, because it is typically not trained independently. Rather, it is derived by fine-tuning the global component or jointly trained with it. The personalized predictor may therefore err on samples that an independently trained model would classify correctly. In cross-device FL, this design is a well-motivated choice that costs little to nothing, since each client typically holds too little data to train a useful model on its own \citep{liu2022privacy,panchal2023flow}. In cross-silo FL, by contrast, clients may hold enough data for an independently trained model to be competitive with one trained collaboratively \citep{liu2022privacy,sun2024collaborate,pye2021personalized}, and such errors may no longer be negligible. 

Regardless of how they address statistical heterogeneity, most of the approaches above also share and aggregate model parameters, requiring all clients to use the same model architecture. In cross-device FL, this requirement is restrictive mainly because of \textbf{system heterogeneity}, i.e., differences in clients' compute and memory, since a shared architecture must satisfy the resource constraints of the least capable devices. In cross-silo FL, where clients are typically organizations with ample computational resources \citep{huang2022crosssilo,wang2021field}, architectural heterogeneity is instead primarily motivated by differences in clients' data and institutional context. Because model families have different inductive biases, their relative performance varies with dataset properties such as sample size and feature distribution \citep{grinsztajn2022why,mcelfresh2023neural}, both of which are prominent sources of statistical heterogeneity across clients in cross-silo FL \citep{kairouz2021advances,huang2022crosssilo}. For instance, hospitals in different regions may serve distinct patient populations and follow different clinical protocols or documentation practices \citep{rieke2020future,thakur2024knowledge,molaei2026learning}. As autonomous data owners, organizations are also generally driven to choose and configure their own models in alignment with institution-specific requirements or priorities
\citep{yang2019federated,rieke2020future,cao2023cofed,xie2024mhpflid,ciupek2025fl}.

These considerations have motivated model-heterogeneous pFL, in which clients maintain personalized models that may differ in architecture, though existing methods vary in the degree of architectural flexibility they support. Some require client models to retain a common sub-component \citep{diao2020heterofl,liang2020think} or to produce representations in a shared embedding space \citep{tan2022fedproto}. Others permit fully distinct architectures without structural constraints, but often still rely on public reference datasets \citep{li2019fedmd,lin2020ensemble,cao2023cofed} or a shared auxiliary model \citep{shen2020federated,qin2023fedapen,xie2024mhpflid}.

In this work, we introduce FedDES (Federated Dynamic Expert Selection), a novel instance-level pFL approach that treats independently trained client models as a pool of experts, with each client learning a graph neural network (GNN) router that dynamically selects and weights a subset of experts for each input. FedDES sits at the intersection of mixture-of-experts (MoE) and dynamic ensemble selection (DES) \citep{cruz2018dynamic}, combining MoE's learned routing mechanism with DES's notion of expert competence, i.e., how likely an expert is to be correct on a given input. Central to this approach is a graph over sample--expert pairs, structured by the similarity among local samples. A GNN propagates information over this graph to learn a joint representation for each sample--expert pair that is predictive of the expert's ability to classify the sample correctly. Each pair's representation is mapped to a competence score, and for each input, the experts' scores are converted into sparse ensemble weights, allowing poorly suited experts to be excluded entirely. 

Preserving individual client models to combine their outputs is a departure from the parameter-aggregation paradigm that dominates federated learning, and is key to how FedDES addresses the limitations discussed above. By aggregating client contributions in the output space, FedDES allows full architectural flexibility. The value of this property for personalization is best understood by considering a personalized predictor that errs on samples that the client's independently trained local model, or another client's model, would classify correctly. An oracle that selected a correct model on exactly those samples would recover every such error, but only if the individual models remain available to select from.
By training the router to concentrate weight on the experts that classify each labeled sample correctly, FedDES aims to approximate the oracle on unseen inputs.

Our contributions are as follows: 
\begin{itemize}[noitemsep,topsep=0pt]
\item \textbf{FedDES.} We introduce FedDES, an instance-level pFL approach that treats clients' heterogeneous models as a pool of experts. Each client trains a graph neural network that learns joint representations of sample--expert pairs, which are used to dynamically select and weight an ensemble of experts for each input.
\item \textbf{Instance-level transfer framework.} We introduce a framework that compares individual predictions of a personalized predictor obtained through collaboration with those that an independently trained local model would produce. A personalized predictor may correctly classify samples on which independent training would err, and may err on samples that independent training would classify correctly. Quantifying how often each occurs explains how a predictor's gain comes about and how methods reach similar gains in different ways.
\item \textbf{Limits of coarse personalization.} Using this framework, we relate the gain a method leaves unrealized to how it pursues personalization. We quantify the gains attainable in principle from personalizing at the instance level rather than the client level, and from selectively integrating knowledge from individual source clients rather than consolidated cross-client knowledge.
\item \textbf{Empirical evaluation.} Across eight data settings with simulated and naturally occurring heterogeneity, FedDES outperforms model-heterogeneous pFL baselines in every setting while most effectively mitigating negative transfer.
\end{itemize}

\section{Related Work}

\subsection{Personalized Federated Learning}

Personalized federated learning aims to learn models tailored to each client's data distribution while retaining the benefits of collaborative learning \citep{kairouz2021advances}. A range of personalization strategies have been developed, including meta-learning approaches that learn shared initializations for rapid local adaptation \citep{fallah2020personalized}, regularization methods that penalize deviation from a global reference \citep{t2020personalized,li2021ditto}, and interpolation between local and global models \citep{deng2020adaptive}. These approaches frame personalization as a balance between local and shared knowledge, which typically comes in the form of a global model. Because individual client contributions are consolidated within the global model, a target client cannot selectively integrate knowledge from relevant clients while rejecting contributions from those whose knowledge is irrelevant or even harmful. Personalized aggregation approaches address this limitation by assigning client-specific weights that reflect the estimated relevance of other clients' models to the target client. FedAMP estimates relevance from model similarity \citep{huang2021personalized}, FedFomo from performance on local validation data \citep{zhang2020personalized}, and APPLE learns the weights directly during training \citep{luo2022apple}. However, these approaches generally rely on parameter sharing or aggregation, requiring clients to use identical or structurally compatible architectures.

Model-heterogeneous pFL relaxes this requirement through several forms of knowledge transfer. Model-splitting methods such as LG-FedAvg \citep{liang2020think} and FedGH \citep{yi2023fedgh} maintain a shared model component while allowing variation in other components, although this constrains architectural choices (e.g., if a global head is shared,
feature extractor output dimensions must match). Prototype-based methods such as FedProto \citep{tan2022fedproto} and FedTGP enable model heterogeneity by communicating class-level feature representations rather than parameters or predictions, but still require client models to produce intermediate feature representations of the same dimension \citep{tan2022fedproto,zhang2024fedtgp}. Knowledge-distillation methods transfer knowledge through model outputs to support fully heterogeneous models, but often rely on a shared reference dataset \citep{li2019fedmd,lin2020ensemble}. Such data can be difficult to obtain, and performance depends on how well it matches the target task distribution. FML and FedKD avoid this dependency through mutual distillation, but must collaboratively train a homogeneous auxiliary model alongside each client's personalized model \citep{shen2020federated,wu2022communication}. Although these model-heterogeneous method families provide architectural freedom (to varying degrees), they generally still aggregate client contributions into a single knowledge source, whether a common model component, shared class prototypes, or an auxiliary model. Once merged, knowledge from individual clients can no longer be selectively incorporated according to its relevance to the target client, as in personalized aggregation.

\subsection{Mixture of Experts and Dynamic Ensemble Selection}
Mixture-of-experts (MoE) and dynamic ensemble selection (DES) provide related foundations for combining predictors based on their relevance to a given input. MoE models combine multiple expert networks through a router (i.e., a gating network) that determines, for each input, how much each expert contributes to the output. Although MoE is now widely used for conditional computation, activating only a few experts per input to scale model capacity efficiently \citep{shazeer2017outrageously,fedus2022switch}, the original formulation was motivated by specialization: by partitioning the input space among experts, each can learn to handle a different region of the problem \citep{jacobs1991adaptive}. In conventional MoE, the experts and router are trained jointly, so expert specialization emerges implicitly from the training objective. DES instead operates on a pool of independently trained classifiers and, for each query, selects a subset based on their performance in the query's vicinity, called the region of competence (RoC) \citep{cruz2018dynamic}. Most DES methods define this region using $k$-nearest neighbors in the feature space \citep{cruz2018dynamic,britto2014dynamic}, but proximity in this space does not necessarily imply shared competent classifiers, particularly in regions of class overlap \citep{oliveira2017online,cruz2017dsvsknn} or in high-dimensional data settings, where distance-based similarity becomes less meaningful \citep{beyer1999nearest}.  Some methods therefore define the region of competence in the decision space, where samples are compared by their output profiles, i.e., the vector of predictions from all experts in the pool \citep{giacinto2001dynamic,cavalin2013dynamic}.  Classical DES methods commonly use accuracy in the RoC as competence estimates, as in KNORA \citep{ko2008dynamic}, whereas meta-learning approaches such as META-DES learn it from multiple indicators of classifier behavior \citep{cruz2015meta}. FedDES draws on this query-specific competence principle, but rather than estimating competence from predefined criteria computed over the RoC, it uses decision-space neighborhoods as a relational inductive bias for a GNN router that learns competence end-to-end.

\subsection{Instance-Level Personalized Federated Learning}

Most pFL approaches perform client-level adaptation. A smaller set of methods have begun to explore instance-level adaptation, and nearly all do so by adaptively combining local and global knowledge for each sample. They primarily differ in how the two are represented and combined. Approaches include weighting the outputs of a global model and a locally fine-tuned copy through a gating network \citep{zec2020specialized,guo2021pflmoe}, mixing the layer-wise outputs of global and local models \citep{panchal2023flow}, dividing a sample's features to be processed separately by a global and a local prediction head \citep{zhang2023fedcp}, and weighting the representations output by a global and a local feature extractor \citep{yi2026pfedmoe}. In these methods, the local component is not trained independently, so even if the personalization mechanism enables clients to rely exclusively on the local component for a given sample, the prediction does not reflect purely local knowledge. In other methods, global knowledge is spread across a set of global experts \citep{reisser2021federated,farhat2025learning}, each collaboratively trained by multiple clients, so that no expert represents the knowledge of an individual client. Regardless of whether global knowledge takes the form of a single shared model or a shared set of models, clients are unable to selectively draw on the knowledge of the source clients most relevant to a given sample.

A notable exception is dFLMoE \citep{xie2025dflmoe}, in which clients exchange the prediction heads of their locally trained models, adapt the outputs of their own feature extractor to each head through a learned transformation, and learn a gating network that weights the heads' predictions for each input. Although each client treats its own and the received heads as individual experts, the predictions of the received heads do not solely reflect knowledge from source clients, since their inputs are produced by the client's own feature extractor and a transformation learned on its local data.

\section{Dynamic Expert Selection}

FedDES operates in a fully decentralized setting and follows the peer-to-peer (P2P) model sharing protocol introduced in \citet{mueller2024fedpae}. Each client carries out three main operations: (1) assembling an expert pool through local training followed by P2P model exchange, (2) evaluating the expert pool on local data to construct a graph over sample--expert pairs, structured by the similarity among local samples, and (3) training a GNN router on this graph to produce sparse, input-dependent routing weights over the experts.

\subsection{Problem Setup}
\label{sec:setup}

We consider $N$ clients, where client $i$ holds a private dataset $\mathcal{D}_i$ of $n_i$ samples drawn from its local data distribution $P_i$ over $\mathcal{X}\times\mathcal{Y}$, where $\mathcal{X}$ denotes the input space and $\mathcal{Y}$ the label space of
$C$ classes. The local data distributions may differ across clients, reflecting statistical heterogeneity. We describe collaboration from the perspective of a single \emph{target client} $i$ and refer to the other clients, whose knowledge it may draw on, as \emph{source clients}.

Each client $i$ independently trains a set $\mathcal{F}_i$ of one or more models on $\mathcal{D}_i$. The models may differ in architecture across clients and need not require gradient-based training. Clients then exchange their trained models so that every client holds the same collection $\mathcal{F} = \bigcup_{i=1}^{N} \mathcal{F}_i$, which we refer to as the expert pool. For simplicity, we focus on the setting in which each client contributes a single model to the pool, as in our main experiments, so that $\mathcal{F} = \{f_i\}_{i=1}^{N}$. For an input $\mathbf{x}$,
expert $f_i$ outputs a probability vector $\mathbf{p}_i(\mathbf{x})$ over the $C$
classes, with predicted class $\hat{y}_i(\mathbf{x}) = \arg\max_c p_{i,c}(\mathbf{x})$.

We also consider intra-client heterogeneity, where a client's data may span distinct
subpopulations. More generally, the relative suitability of experts may vary across
individual inputs, even without identifiable subpopulations.

To accommodate this variation, each client $i$ learns a router,
parameterized by $\phi_i$, that selects and weights a subset of experts
for each input $\mathbf{x}$. The resulting sparse weights 
$\boldsymbol{\alpha}_i(\mathbf{x})\in\Delta^{N-1}$ yield the personalized predictor
$$
\mathbf{g}_i(\mathbf{x}) = \sum_{j=1}^{N}\alpha_{ij}(\mathbf{x})\,\mathbf{p}_j(\mathbf{x}),
$$
a convex combination of the experts' outputs, where $\Delta^{N-1}$ denotes the
probability simplex over the $N$ experts and $\alpha_{ij}(\mathbf{x}) = 0$ indicates
that expert $j$ is excluded from the prediction. The corresponding predicted class is
$\hat{y}_{\mathbf{g}_i}(\mathbf{x})=\arg\max_c g_{i,c}(\mathbf{x})$. Given a classification loss $\ell$, the goal of client $i$ is to solve
$$
\min_{\phi_i} \; \mathbb{E}_{(\mathbf{x},y)\sim P_i}
\left[ \ell\big(\mathbf{g}_i(\mathbf{x}),y\big) \right].
$$

Beyond minimizing predictive loss, we aim to improve each client's performance
relative to independent local training. Let $\operatorname{Acc}_i(h)$
denote the expected accuracy of predictor $h$ on client $i$'s distribution. We
define the accuracy gain from collaboration for client $i$ as
$$
\Delta_i = \operatorname{Acc}_i(\mathbf{g}_i) - \operatorname{Acc}_i(f_i).
$$
Client-level negative transfer occurs when $\Delta_i < 0$, that is, when collaboration lowers client $i$'s overall accuracy below that of its local model $f_i$. This is the sense in which the term is typically used in pFL. The pFL literature commonly reports performance averaged across clients
\citep{chen2022pflbench,wu2022motley}, which can conceal client-level negative
transfer \citep{yu2020salvaging,li2020fair,cho2022federate}.

Two methods can attain the same $\Delta_i$ while having very different prediction patterns relative to the local model. One may rarely disagree with the local model, while another may correctly classify many samples that the local model misclassifies, offset by misclassifying many that the local model classifies correctly. Examining these patterns can help explain how different personalization strategies improve on local training or fall short of it, and where gains are left unrealized. We therefore introduce an instance-level transfer framework that compares the predictions of the personalized predictor and the local model on individual samples. The local model serves as the reference because it represents the counterfactual in which client $i$ does not participate in collaborative learning. Following the terminology used in the model-update literature \citep{yan2021positive}, we define a \emph{negative flip} as the case where a sample is correctly classified by $f_i$ while being misclassified by $\mathbf{g}_i$, and a \emph{positive flip} as the case where a sample is misclassified by $f_i$ while being correctly classified by $\mathbf{g}_i$. For client $i$, the corresponding rates are

$$
\mathrm{PFR}_i = \Pr_{(\mathbf{x},y)\sim P_i}
\left[ \hat{y}_i(\mathbf{x})\neq y,\; \hat{y}_{\mathbf{g}_i}(\mathbf{x})= y \right],
\qquad
\mathrm{NFR}_i = \Pr_{(\mathbf{x},y)\sim P_i}
\left[ \hat{y}_i(\mathbf{x})=y,\; \hat{y}_{\mathbf{g}_i}(\mathbf{x})\neq y \right].
$$
The client-level gain decomposes exactly as
$$
\Delta_i = \mathrm{PFR}_i - \mathrm{NFR}_i,
$$
which makes precise the distinction above. To our knowledge, prior work in pFL has not examined a method's gain over local training through the lens of the individual predictions on which the two differ. These definitions apply to any personalized predictor in place of $\mathbf{g}_i$, which is how we use them to compare methods in Section~\ref{sec:experiments}.

A nonzero $\mathrm{NFR}_i$ is not in itself an argument against collaboration, nor a quantity to minimize on its own, since a predictor that always returns $f_i(\mathbf{x})$ attains $\mathrm{NFR}_i = \mathrm{PFR}_i = 0$. Rather, it measures gain that the predictor leaves unrealized. Consider the oracle that returns $\mathbf{g}_i(\mathbf{x})$, except on the samples where $f_i$ is correct and $\mathbf{g}_i$ is not, where it returns $f_i(\mathbf{x})$ instead. Its accuracy is
$$
\operatorname{Acc}_i(\mathbf{g}_i) + \mathrm{NFR}_i,
$$
so $\mathrm{NFR}_i$ is exactly the accuracy gap between the personalized predictor $\mathbf{g}_i$ and this oracle. 

We therefore read the two rates together, as a description of how a predictor's gain is composed, rather than treating either one as a measure of performance.

The oracle above uses only $f_i$ and $\mathbf{g}_i$, so its accuracy bounds what a per-sample choice between local and personalized predictions can achieve. A second oracle bounds what per-sample selection over a pool consisting of all clients' independently trained models can achieve. On each sample, it returns the prediction of a model in $\mathcal{F}$ that classifies the sample correctly, whenever one exists, so a predictor's unrealized gain up to this ceiling is
$$
\operatorname{Acc}_i(\text{pool oracle}) - \operatorname{Acc}_i(\mathbf{g}_i)
= \mathrm{NFR}_i + \mathrm{MFR}_i - \mathrm{BPR}_i,
$$
where
$$
\mathrm{MFR}_i = \Pr_{(\mathbf{x},y)\sim P_i}\!\left[\hat{y}_i(\mathbf{x}) \neq y,\;
\hat{y}_{\mathbf{g}_i}(\mathbf{x}) \neq y,\;
\exists\, j:\ \hat{y}_j(\mathbf{x}) = y\right],
\qquad
\mathrm{BPR}_i = \Pr_{(\mathbf{x},y)\sim P_i}\!\left[\hat{y}_{\mathbf{g}_i}(\mathbf{x}) = y,\;
\forall\, j:\ \hat{y}_j(\mathbf{x}) \neq y\right].
$$
$\mathrm{MFR}_i$, the \emph{missed flip rate}, is the share of samples on which neither the local model nor the personalized predictor is correct but some model in $\mathcal{F}$ is. $\mathrm{BPR}_i$, the \emph{beyond-pool rate}, is the share of samples that the personalized predictor classifies correctly while no model in $\mathcal{F}$ does. For a predictor that combines the outputs of the models
in $\mathcal{F}$, as in FedDES, beyond-pool gains are attainable only through ensemble predictions that are correct despite every individual model being wrong. For a predictor whose model is trained jointly with other clients, as in the other pFL methods we evaluate, beyond-pool gains arise from correctly classifying samples that no independently trained model classifies correctly.

For a predictor obtained through client-level personalization that draws on consolidated cross-client knowledge, $\mathrm{NFR}_i$ quantifies the unrealized gain recoverable
through instance-level selection between local and personalized
predictions, whereas $\mathrm{MFR}_i$ quantifies the additional
gain recoverable by selecting among independently trained
client models. Since methods that consolidate cross-client knowledge do not retain individual source models for selection at inference,
the pool oracle serves as a reference point rather than a
ceiling for such methods. We report these quantities for
every method on that basis.

Sections~\ref{sec:graph}--\ref{sec:training} describe how each client constructs its graph and trains its router. Since these steps are carried out independently by each client, we describe them for a single client and omit the client index $i$.

\subsection{Graph Construction}
\label{sec:graph}

\begin{figure}[t]
  \centering
  \includegraphics[width=\linewidth]{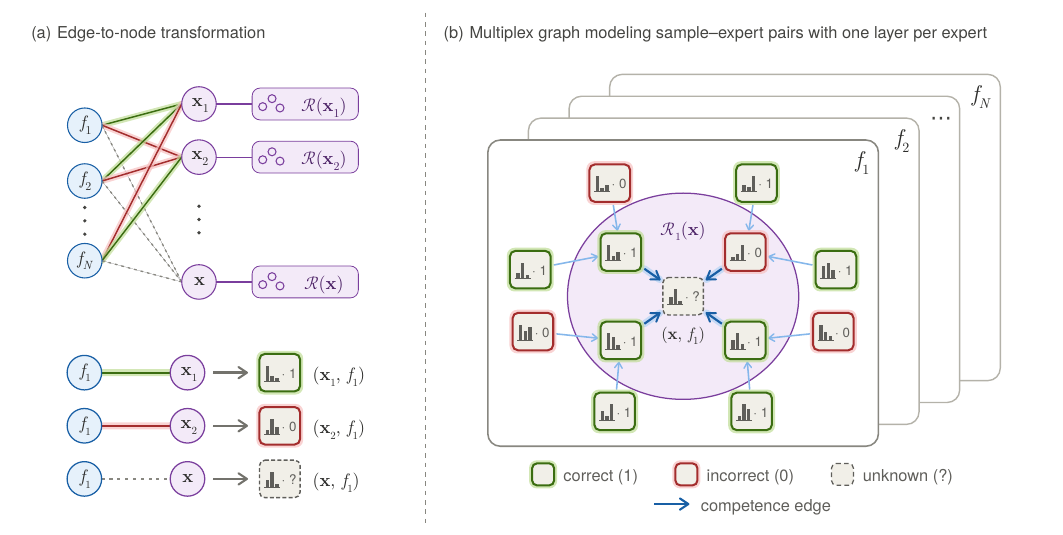}
  \caption{\textbf{Graph construction in FedDES.}
  (a) Conceptual heterogeneous graph in which each expert is linked to every local sample by an edge carrying its prediction and, for labeled samples, whether the expert is correct. Each sample is also linked to its nearest training samples in decision space, which form its region of competence $\mathcal{R}(\mathbf{x})$. An edge-to-node transformation turns each edge with endpoints $\mathbf{x}$ and $f_j$ into a node $(\mathbf{x}, f_j)$ representing the pair. Its features are the expert's predicted probability vector $\mathbf{p}_j(\mathbf{x})$, shown as a histogram, and its correctness status, marked 1 (\textsc{correct}), 0 (\textsc{incorrect}), or ? (\textsc{unknown}).
  (b) The nodes of each expert form one layer of a multiplex graph, structured by regions of competence. In layer $f_1$, the unseen node $(\mathbf{x}, f_1)$ of a test sample $\mathbf{x}$ receives competence edges from the nodes in $\mathcal{R}_1(\mathbf{x}) = \{(\mathbf{x}', f_1) : \mathbf{x}' \in \mathcal{R}(\mathbf{x})\}$, each of which has its own set of neighbors defined in the same way.}
  \label{fig:graph}
\end{figure}

FedDES approaches instance-level expert selection by modeling the relationships between local samples and experts. A natural way to represent these relationships is a heterogeneous graph with sample nodes and expert nodes, in which samples are linked by similarity and each expert is linked to every sample by an edge carrying its prediction (Figure~\ref{fig:graph}a). Under this design, a GNN learns representations for all sample and expert nodes, then scores each expert against each sample. However, evidence about expert $f_j$ on a sample $\mathbf{x}'$ reaches a similar sample $\mathbf{x}$ only through the node of $f_j$, whose representation summarizes the behavior of $f_j$ on every sample, or through the node of $\mathbf{x}'$, whose representation summarizes the behavior of every expert on $\mathbf{x}'$. FedDES instead builds a graph in which each sample--expert pair $(\mathbf{x}, f_j)$ is a node with its own representation, and samples and experts serve only to index these nodes (Figure~\ref{fig:graph}b).  Two nodes are connected when the corresponding edges share an expert and have samples that are linked in the conceptual heterogeneous graph  (Figure~\ref{fig:graph}a). The nodes of each expert therefore form one layer of a multiplex graph \citep{kivela2014multilayer}, with edges structured by sample similarity. Constructions of this kind are used in GNNs to learn edge representations directly \citep{jiang2019censnet,cai2022line}. Evidence about $f_j$ on $\mathbf{x}'$ then passes directly from node $(\mathbf{x}', f_j)$ to node $(\mathbf{x}, f_j)$. 

\paragraph{Nodes.} The graph contains one node $(\mathbf{x}, f_j)$ for each pair of a local sample $\mathbf{x}$ and an expert $f_j$. We denote the nodes of sample $\mathbf{x}$, one in each multiplex layer, by $\mathcal{V}(\mathbf{x}) = \{(\mathbf{x}, f_j)\}_{j=1}^{N}$. The features of node $(\mathbf{x}, f_j)$ describe the predictive behavior of $f_j$ on $\mathbf{x}$. Its feature vector is the concatenation of the probability vector $\mathbf{p}_j(\mathbf{x})$ and a one-hot correctness status $s_j(\mathbf{x})$. For a training sample, $s_j(\mathbf{x})$ encodes \textsc{correct} or \textsc{incorrect} according to whether the predicted class matches the ground-truth label. The status takes a third value, \textsc{unknown}, for unseen nodes, described later in this section, and for nodes masked during training (Section~\ref{sec:training}).

\paragraph{Competence edges.}
Competence edges connect nodes from the same expert and carry evidence about how that expert performed on samples similar to a given input. We measure similarity between samples in the decision space \citep{giacinto2001dynamic,cavalin2013dynamic}. Each sample $\mathbf{x}$ is mapped to this space through its output profile, the concatenation of the probability vectors of all experts,
$$
\mathbf{o}(\mathbf{x}) = \big[\mathbf{p}_1(\mathbf{x}), \ldots, \mathbf{p}_N(\mathbf{x})\big] \in \mathbb{R}^{NC}.
$$
Whereas distances between raw features can largely reflect low-level variation for some modalities, such as differences in pixel intensities between images, distances between output profiles provide a similarity measure that applies consistently across data modalities. For each sample $\mathbf{x}$, let $\mathcal{R}(\mathbf{x})$ denote its region of competence, consisting of its $k$ nearest samples in the training set $\mathcal{D}^{\mathrm{tr}}$ under the $\ell_1$ distance between output profiles, $\delta(\mathbf{x}', \mathbf{x})$, excluding $\mathbf{x}$ itself.\footnote{Because the client's local expert is trained on $\mathcal{D}^{\mathrm{tr}}$, its predictions on these samples are optimistically biased. Following stacked generalization \citep{wolpert1992stacked}, we instead use out-of-fold predictions from 5-fold cross-validation, in which each training sample is predicted by a model of the same architecture trained on the remaining folds. The expert applied to validation and test samples, and shared with other clients, is trained on all of $\mathcal{D}^{\mathrm{tr}}$.} Each node $(\mathbf{x}, f_j)$ then has incoming competence edges from $(\mathbf{x}', f_j)$ for each $\mathbf{x}' \in \mathcal{R}(\mathbf{x})$, weighted by $w(\mathbf{x}', \mathbf{x}) = \operatorname{softmax}_{\mathbf{x}' \in \mathcal{R}(\mathbf{x})}\big(-\delta(\mathbf{x}', \mathbf{x}) / \tau_{\mathbf{x}}\big)$. The temperature $\tau_{\mathbf{x}}$ is the median distance from $\mathbf{x}$ to the samples in $\mathcal{R}(\mathbf{x})$, which adapts the weighting to the local density around $\mathbf{x}$ \citep{zelnik2004self}.


\paragraph{Unseen nodes.}
A validation or test sample $\mathbf{x}$ is added to the graph as a set of unseen nodes $\mathcal{V}(\mathbf{x})$, one in each multiplex layer. Each expert is first evaluated on $\mathbf{x}$, and the resulting predictions form the features of the nodes in $\mathcal{V}(\mathbf{x})$, with the correctness status set to \textsc{unknown}. These predictions also define the output profile of $\mathbf{x}$, from which its region of competence $\mathcal{R}(\mathbf{x})$ is identified among the training samples. Each node $(\mathbf{x}, f_j)$ then has incoming competence edges from $(\mathbf{x}', f_j)$ for each $\mathbf{x}' \in \mathcal{R}(\mathbf{x})$. Because competence edges always originate at nodes of training samples, information from $\mathcal{V}(\mathbf{x})$ cannot reach any other node, including other unseen nodes.

\subsection{Router}
\label{sec:router}

\begin{figure}[t]
  \centering
  \includegraphics[width=\linewidth]{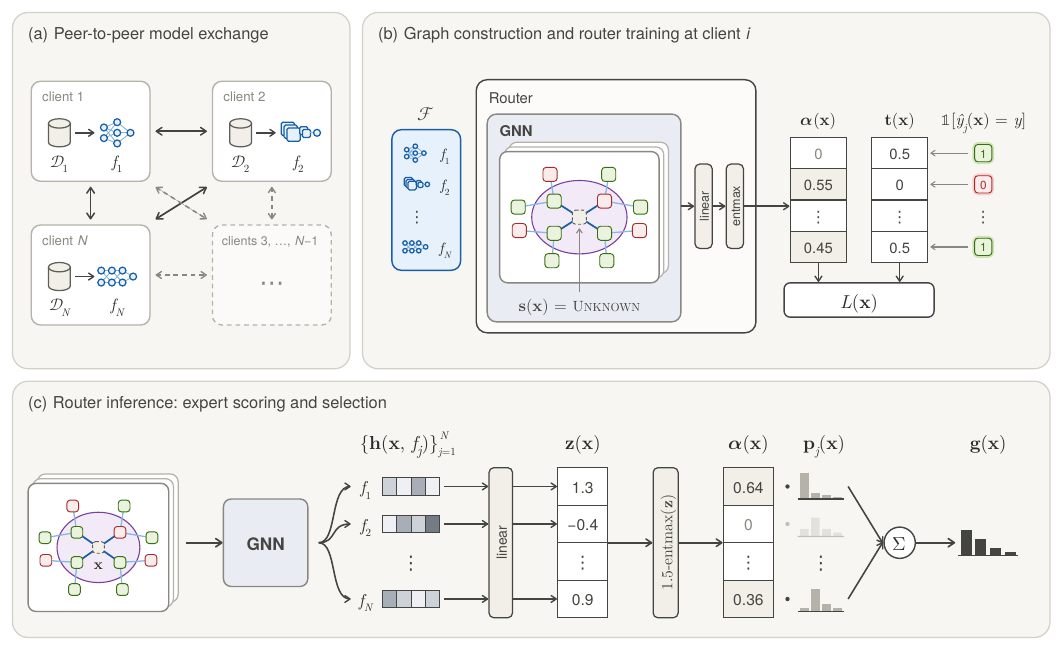}
  \caption{\textbf{Overview of FedDES.}
  (a) Each client trains a model on its local data, and clients exchange models peer to peer so that every client holds the same expert pool $\mathcal{F}$.
  (b) At client $i$, the pool is evaluated on the local training samples to build a graph over sample--expert pairs (Figure~\ref{fig:graph}). During training, the correctness status of a masked sample $\mathbf{x}$ is set to \textsc{unknown}. The router produces ensemble weights $\boldsymbol{\alpha}(\mathbf{x})$, which the loss $L(\mathbf{x})$ compares to a target distribution $\mathbf{t}(\mathbf{x})$ that is uniform over the experts that classify $\mathbf{x}$ correctly, that is, those with $\ind[\hat{y}_j(\mathbf{x}) = y] = 1$. In the example shown, $L(\mathbf{x})$ therefore encourages the router to assign less weight to $f_2$ and more to $f_1$.
  (c) A test sample $\mathbf{x}$ is added to the graph as a set of unseen nodes $\mathcal{V}(\mathbf{x})$. The GNN produces their embeddings $\{\mathbf{h}(\mathbf{x}, f_j)\}_{j=1}^{N}$, a linear readout maps them to competence scores $\mathbf{z}(\mathbf{x})$, and entmax converts the scores into ensemble weights $\boldsymbol{\alpha}(\mathbf{x})$. The prediction $\mathbf{g}(\mathbf{x})$ is the weighted sum of the selected experts' probability vectors $\mathbf{p}_j(\mathbf{x})$.}
  \label{fig:overview}
\end{figure}

Given the graph described above, each client trains a GNN-based router to assign each node $(\mathbf{x}, f_j)$ a competence score reflecting how likely $f_j$ is to classify $\mathbf{x}$ correctly. These scores are mapped to ensemble weights $\boldsymbol{\alpha}(\mathbf{x})$ that determine which experts contribute to the prediction and in what proportion.

\paragraph{Message passing.}
The GNN consists of $T$ layers, each with its own parameters. The feature vector of each node is linearly projected to obtain its
initial representation $\mathbf{h}^{(0)}(\mathbf{x}, f_j) \in \mathbb{R}^d$, where $d$
is the GNN's hidden dimension. At layer
$t \in \{1, \dots, T\}$, every node is updated using the previous layer's
representations, which we write as $\mathbf{h}(\cdot) = \mathbf{h}^{(t-1)}(\cdot)$.

Along \emph{competence edges}, node $(\mathbf{x}, f_j)$ attends over $(\mathbf{x}', f_j)$ for $\mathbf{x}' \in \mathcal{R}(\mathbf{x})$, using the graph transformer convolution of \citet{shi2021masked} with $w(\mathbf{x}', \mathbf{x})$ as an edge feature, which yields the message $\mathbf{m}^{\mathrm{comp}}(\mathbf{x}, f_j)$. This message conveys how expert $f_j$ behaved on similar training samples. In the first layer, the message captures information similar to that used by DES methods that estimate competence from a classifier's performance on samples resembling the input \citep{cruz2018dynamic}. In this sense, the neighborhood structure serves as an inductive bias, encoding the locality assumption of DES, but the router is able to learn how much to rely on local performance and, through attention, which neighbors are most informative about $f_j$'s competence on $\mathbf{x}$.

The resulting message is combined with the node's representation through a residual connection and passed through a feed-forward network. Full architectural details are given in Appendix~\ref{app:arch}.

\paragraph{Expert scoring and selection.}

The final layer produces, for each node $(\mathbf{x}, f_j)$, an embedding $\mathbf{h}^{(T)}(\mathbf{x}, f_j)$. Whereas the node's input features describe only expert $f_j$'s prediction on $\mathbf{x}$, its embedding encodes relational evidence, such as how $f_j$ performs on samples similar to $\mathbf{x}$. A linear readout maps each embedding to a competence score $z_j(\mathbf{x})$. The scores of all experts on $\mathbf{x}$ are then passed to entmax \citep{peters2019sparse}, a sparse alternative to softmax, which converts these scores into ensemble weights (Figure~\ref{fig:overview}c),
$$
\boldsymbol{\alpha}(\mathbf{x}) = \operatorname{entmax}\big(z_1(\mathbf{x}), \ldots, z_N(\mathbf{x})\big).
$$
Unlike softmax, entmax can assign exactly zero weight to low-scoring experts, so each sample is predicted by its own subset of experts, of variable size, through the personalized predictor $\mathbf{g}(\mathbf{x})$ of Section~\ref{sec:setup}.

Because the features and connections of unseen nodes are computed from the experts' predictions alone, and no GNN parameters are tied to a particular sample, the model is inductive, meaning that it can score unseen nodes without retraining. Likewise, no node feature contains an expert identifier and no parameters are tied to a particular expert. As a result, the GNN architecture does not depend on the composition or size of the expert pool, allowing new clients to join the federation without architectural changes.

\subsection{Training}
\label{sec:training}

Since the correctness of the experts on a test sample is unknown at inference, we reproduce this condition during training by masking. In each epoch, training samples are independently selected
for masking with probability $\rho$. Let
$\mathcal{M} \subset \mathcal{D}^{\mathrm{tr}}$ denote the subset
of masked samples. For all $\mathbf{x} \in \mathcal{M}$,
the nodes $\mathcal{V}(\mathbf{x})$ are treated as unseen nodes. The correctness status of these nodes is set to \textsc{unknown}, all their incoming and outgoing competence edges are removed, and each node $(\mathbf{x}, f_j)$ in $\mathcal{V}(\mathbf{x})$ receives new incoming competence edges from $(\mathbf{x}', f_j)$ for the $k$ nearest samples $\mathbf{x}' \in \mathcal{D}^{\mathrm{tr}} \setminus \mathcal{M}$, as for unseen nodes (Section~\ref{sec:graph}). Unmasked nodes that lose their incoming competence edges from masked nodes for that epoch receive no replacements, and their remaining edges retain their original weights. The router is trained to predict expert competence on the masked samples only, so each score is learned from the same information available at an unseen node (Figure~\ref{fig:overview}b).

For a training sample $(\mathbf{x}, y)$, let
$\mathcal{J}(\mathbf{x}) = \{ j : \hat{y}_j(\mathbf{x}) = y \}$ denote the set of experts
that classify it correctly. The router is trained to place equal weight on these
experts, with target
$$
t_j(\mathbf{x}) = \frac{\ind\big[j \in \mathcal{J}(\mathbf{x})\big]}{|\mathcal{J}(\mathbf{x})|}.
$$

The router is trained with the Fenchel--Young loss associated with entmax
\citep{blondel2020learning}, which directly optimizes the weights produced at
prediction time. For a sample $\mathbf{x}$, let
$\mathbf{z}(\mathbf{x}) = \big(z_1(\mathbf{x}), \dots, z_N(\mathbf{x})\big)$ denote the
experts' competence scores and
$\boldsymbol{\alpha}(\mathbf{x}) = \operatorname{entmax}\big(\mathbf{z}(\mathbf{x})\big)$
the resulting ensemble weights. The loss is
$$
L(\mathbf{x}) =
\big(\boldsymbol{\alpha}(\mathbf{x}) - \mathbf{t}(\mathbf{x})\big)^{\top}\mathbf{z}(\mathbf{x})
+ \mathsf{H}\big(\boldsymbol{\alpha}(\mathbf{x})\big)
- \mathsf{H}\big(\mathbf{t}(\mathbf{x})\big),
$$
where $\mathsf{H}$ is the Tsallis entropy of order 1.5, a generalization of the Shannon
entropy whose order controls the sparsity of the entmax weights. With the Shannon
entropy in its place, entmax reduces to softmax. The loss reaches its minimum of zero when the weights match the target. Because its gradient with respect to the scores is
$\boldsymbol{\alpha}(\mathbf{x}) - \mathbf{t}(\mathbf{x})$, $L(\mathbf{x})$ encourages the router to lower the score of each incorrect expert in proportion to its weight and to rebalance the correct experts toward equal shares. Incorrect experts already assigned zero weight receive no
gradient. The loss is averaged over the masked samples on which some, but not all, experts are
correct, $0 < |\mathcal{J}(\mathbf{x})| < N$. When every expert is correct, any weighting yields a correct prediction, and when every expert is incorrect, the target is undefined.

\begin{table}[b]
\centering
\small
\caption{Summary of datasets, client partitions, and evaluation metrics.}
\label{tab:datasets}
\vspace{4pt}
\begin{tabular*}{\textwidth}{@{\extracolsep{\fill}}llrrrrccl}
\toprule
 & & & \multicolumn{2}{c}{Total} & \multicolumn{3}{c}{Per client} & \\
\cmidrule(lr){4-5} \cmidrule(lr){6-8}
Dataset & Partitioned by & $N$ & Samples & Classes & Samples & Classes & Positive & Metric \\
\midrule
CIFAR-10 & label, Dir($\alpha=0.5$) & 20 & 60K & 10 & 1.4K--5.0K & 3--10 & -- & Acc. \\
CIFAR-10 & label, Dir($\alpha=0.3$) & 20 & 60K & 10 & 0.5K--6.3K & 3--10 & -- & Acc. \\
CIFAR-10 & label, Dir($\alpha=0.1$) & 20 & 60K & 10 & 0.3K--7.4K & 1--10 & -- & Acc. \\
DomainNet & domain, Dir($\alpha=0.5$) & 12 & 22K & 10 & 0.6K--3.7K & 10 & -- & Acc. \\
BraTS & institution & 8 & 74K & 2 & 1.9K--34.1K & -- & 48--55\% & Acc. \\
eICU & hospital & 50 & 52K & 2 & 0.3K--3.1K & -- & 6--23\% & BA \\
scRNA-seq & study & 4 & 14K & 8 & 1.4K--8.5K & 3--8 & -- & BA \\
PaySim & simulated bank & 5 & 1.5M & 2 & 300K & -- & 0.1--0.2\% & AUPRC \\
\bottomrule
\end{tabular*}

\vspace{4pt}
\begin{minipage}{\textwidth}
\footnotesize
Samples include each client's training, validation, and test data. Per-client values give the smallest and largest value over all clients in the three seeds, with the classes present shown for multiclass tasks and the positive rate for binary tasks. Acc.: accuracy; BA: balanced accuracy.
\end{minipage}
\end{table}

\section{Experiments}

\subsection{Experimental Setup}
\label{sec:experiments}

\paragraph{Datasets and partitioning.} We evaluate FedDES on two image benchmarks with simulated heterogeneity, three real-world biomedical datasets with naturally occurring heterogeneity, and a simulated financial fraud dataset. Each client's data is split into training (70\%), validation (15\%), and test (15\%) sets, stratified by label. Dataset statistics are summarized in Table~\ref{tab:datasets}, and preprocessing details are provided in Appendix~\ref{app:datasets}.

\textit{CIFAR-10.} CIFAR-10~\citep{krizhevsky2009learning} is a widely used image classification benchmark, consisting of 60,000 images across 10 classes. We simulate a federation of 20 clients by sampling each client's class proportions from a Dirichlet distribution \citep{hsu2019measuring} with $\alpha \in \{0.5, 0.3, 0.1\}$, where smaller values of $\alpha$ correspond to greater heterogeneity across clients.

\textit{DomainNet.} DomainNet~\citep{peng2019moment} is an image classification benchmark spanning six visual domains (clipart, infograph, painting, quickdraw, real, and sketch). Following FedBN~\citep{li2021fedbn}, we use a 10-class subset (22,426 images). We simulate a federation of 12 clients by sampling each client's domain proportions from a Dirichlet distribution with $\alpha = 0.5$, so that clients differ mainly in image style, with each client drawing most of its images from one or two domains.

\textit{BraTS.} The Brain Tumor Segmentation (BraTS) dataset~\citep{menze2015brats, bakas2017advancing} contains multi-parametric MRI scans collected from multiple institutions. We use the FeTS 2022 data partition~\citep{pati2022fets}, treating each institution as a client, and formulate tumor detection as predicting whether a 2D axial slice contains tumor. We retain institutions with at least 25 subjects, yielding $N=8$ clients. Because each subject contributes many slices, each client's data is split by subject, so that all slices from a subject fall in the same split.

\textit{eICU.} The eICU Collaborative Research Database~\citep{pollard2018eicu} contains de-identified electronic health records from ICU admissions across over 200 U.S. hospitals. We consider the binary task of predicting in-hospital mortality from the first 48 hours of each ICU stay. Inputs are hourly time series over the observation window, and static features that cover demographics, admission characteristics, medical history, and active diagnoses. Given the class imbalance, we retain the 50 hospitals with the most positive cases as clients, so that each client's test set contains enough positive cases for stable evaluation. For each client dataset, we perform group-aware splitting on patient ID to prevent data leakage from patients with multiple stays.

\textit{scRNA-seq.} We consider cell-type annotation from single-cell RNA sequencing data using four human pancreas studies~\citep{baron2016, muraro2016, segerstolpe2016, xin2016}, treating each study as a client ($N=4$). Because the studies differ in sequencing protocol and donor population, this dataset exhibits feature-level heterogeneity (i.e., batch effects) in addition to differences in cell-type composition.

\textit{PaySim.} PaySim~\citep{lopezrojas2016paysim} is a synthetic mobile-money transaction dataset for fraud detection, generated by an agent-based simulator calibrated on real transaction logs. We use Flower's federated version~\citep{flower-paysim-banks}, which partitions accounts into five simulated banks with differing fraud rates, treating each bank as a client ($N=5$). We randomly sample 300,000 transactions per bank and remove account-balance features, which cause label leakage through simulator artifacts.

\paragraph{Baselines and models.} Because architectural flexibility is central to FedDES's design, we compare against five MHPFL methods selected from relevant method families: FedGH~\citep{yi2023fedgh} (model splitting), FML~\citep{shen2020federated} (mutual distillation with a proxy model), FedTGP~\citep{zhang2024fedtgp} (prototype-based aggregation), and two instance-level methods, pFedMoE~\citep{yi2026pfedmoe}, which combines local and global knowledge for each sample, and dFLMoE~\citep{xie2025dflmoe}, which weights individual clients' prediction heads for each sample. Two reference baselines are also included: Local, where each client trains its own model independently, and Global Ensemble, where all locally trained models making up the expert pool are weighted equally. 

For each dataset, we use $K$ different model architectures, assigned to clients in round-robin order following \citet{zhang2024fedtgp} (Table~\ref{tab:models}).  FedGH, FedTGP, and pFedMoE all require client models to produce features of a common dimension: FedGH trains a shared prediction head on all clients' representations, FedTGP aggregates class prototypes computed from them, and pFedMoE combines each client's representation with that of a shared feature extractor. Following the original papers, we append a linear projection to each client model for FedGH and pFedMoE, and an average pooling layer for FedTGP, mapping features to the output dimension of the shared architecture in Table~\ref{tab:models}. While dFLMoE clients may also receive prediction heads whose input dimension differs from the output dimension of their feature extractor, no such adaptation is needed, since learned transformations map the feature extractor outputs to each head's input dimension. FML and pFedMoE additionally require a homogeneous auxiliary model shared by all clients, for which we also use the shared architecture. Architecture details are provided in Appendix~\ref{app:models}.

\begin{table}[!ht]
\caption{Client model architectures. The CNN and TPC-light are adapted from the cited architectures (Appendix~\ref{app:models}). Shared: the auxiliary model trained by FML and pFedMoE.}
\label{tab:models}
\footnotesize
\begin{tabular*}{\textwidth}{@{\extracolsep{\fill}} l llll @{}}
\toprule
 & CIFAR-10, DomainNet, BraTS & eICU & scRNA-seq & PaySim \\
\midrule
Architectures & CNN & GRU & MLP, 1 hidden layer & MLP, 2 hidden layers \\
 & \citep{mcmahan2017communication} & \citep{cho2014learning} & & \\[3pt]
 & ResNet-18 & TPC-light & Gated MLP & DCN-V2 \\
 & \citep{he2016deep} & \citep{rocheteau2021temporal} & \citep{dauphin2017language} & \citep{wang2021dcnv2} \\[3pt]
 & MobileNetV2 & Feed-forward attention & & TabNet \\
 & \citep{sandler2018mobilenetv2} & \citep{raffel2015feedforward} & & \citep{arik2021tabnet} \\
\midrule
Shared & CNN & GRU & MLP & DCN-V2 \\
\bottomrule
\end{tabular*}
\end{table}

\paragraph{Implementation details.} We use SGD as the default optimizer, with a learning rate of 0.01, lowered for the few method–dataset combinations that diverged at this value (Appendix~\ref{app:hyperparams}), and a batch size of 32, increased to 1,024 for PaySim given the size of its client datasets. Training runs for up to 500 communication rounds, with full client participation and one epoch of local training per round. To address class imbalance in eICU, PaySim, and scRNA-seq, we weight the training loss by inverse class frequency. For scRNA-seq, class frequencies are computed from the pooled training data, since some clients lack some cell types. Early stopping is applied with a patience of 50 rounds based on average validation performance across clients, except for Local and FedDES's experts, which use each client's own validation performance. For each client, we report test performance at its best-validation round. All methods are implemented in PyTorch within a common framework and run on NVIDIA GeForce RTX 2080 Ti and NVIDIA A40 GPUs.

A small number of method-specific hyperparameters are selected by average validation performance across clients. The prototype loss weight $\lambda \in \{0.01, 0.1, 1\}$ of FedTGP and the learning rate of dFLMoE's feature transforms and router, chosen from $\{10^{-4}, 10^{-3}, 10^{-2}\}$, are tuned on the first seed and reused for the remaining two seeds. For FedDES, the number of neighbors $k \in \{7, 15, 25\}$ and hidden dimension $d \in \{128, 256\}$ are selected for each dataset on the first seed and reused for the remaining two seeds. Other method-specific hyperparameters follow the original papers where reported. Appendix~\ref{app:hyperparams} lists all values, along with the rationale for those not reported in the original papers.

\textbf{Evaluation.} The evaluation metric for each dataset is listed in Table~\ref{tab:datasets}: accuracy for CIFAR-10, BraTS, and DomainNet, balanced accuracy for eICU and scRNA-seq given their class imbalance, and AUPRC for PaySim, where positives are under 0.2\% of transactions. We report the mean $\pm$ standard deviation over three seeded runs, where each seed controls model initialization, data shuffling, and data splitting (train/validation/test sets and, for datasets with simulated heterogeneity, the partition across clients). We consider negative transfer to occur when a client's test performance under collaborative learning falls short of Local by more than 0.5 points, a margin that absorbs seed-level noise in individual client scores. We define the negative transfer rate (NTR) as the fraction of clients that experience negative transfer, and the negative transfer burden (NTB) as the total performance decrease relative to Local incurred by those clients, divided by the total number of clients, so that it reflects both prevalence and severity. Both NTR and NTB are averaged over seeds. We also apply the instance-level transfer framework of Section~\ref{sec:setup}. Because these quantities decompose changes in accuracy, we restrict this analysis to CIFAR-10, BraTS, and DomainNet, where accuracy is also the metric on which methods are compared.

\subsection{Performance Comparison}

\begin{table}[b]
\centering
\caption{Test performance (mean $\pm$ SD over three seeds), with each client evaluated at its best-validation round. Metrics are listed in Table~\ref{tab:datasets}. Bold marks the best method within each data setting.}
\label{tab:main}
\resizebox{\textwidth}{!}{%
\begin{tabular}{l cccccccc}
\toprule
& \multicolumn{3}{c}{CIFAR-10} & & & & & \\
\cmidrule(lr){2-4}
Method & Dir(0.1) & Dir(0.3) & Dir(0.5) & DomainNet & BraTS & eICU & scRNA-seq & PaySim \\
\midrule
Local & 87.3 $\pm$ 2.6 & 72.8 $\pm$ 2.5 & 68.5 $\pm$ 2.3 & 46.2 $\pm$ 2.1 & 87.0 $\pm$ 0.8 & 66.2 $\pm$ 1.0 & 97.4 $\pm$ 0.4 & 23.7 $\pm$ 4.0 \\
FedGH & 87.4 $\pm$ 2.9 & 73.1 $\pm$ 2.3 & 67.2 $\pm$ 2.3 & 44.5 $\pm$ 2.2 & 87.8 $\pm$ 1.1 & 66.7 $\pm$ 0.5 & 97.9 $\pm$ 0.3 & 23.2 $\pm$ 4.3 \\
FML & 88.3 $\pm$ 2.8 & 75.3 $\pm$ 1.8 & 71.1 $\pm$ 2.5 & 50.2 $\pm$ 0.5 & 88.0 $\pm$ 0.1 & 68.1 $\pm$ 0.2 & 97.6 $\pm$ 0.6 & 22.4 $\pm$ 5.7 \\
FedTGP & 88.3 $\pm$ 2.6 & 75.1 $\pm$ 2.3 & 69.9 $\pm$ 2.7 & 47.3 $\pm$ 0.8 & 86.2 $\pm$ 0.4 & 61.0 $\pm$ 0.7 & 97.8 $\pm$ 0.7 & 19.8 $\pm$ 5.2 \\
pFedMoE & 86.7 $\pm$ 2.6 & 72.0 $\pm$ 2.2 & 67.3 $\pm$ 2.5 & 45.5 $\pm$ 1.1 & 83.1 $\pm$ 0.5 & 66.6 $\pm$ 1.0 & 96.7 $\pm$ 1.2 & 23.3 $\pm$ 4.9 \\
dFLMoE & 87.6 $\pm$ 2.7 & 73.3 $\pm$ 2.8 & 68.1 $\pm$ 2.3 & 46.3 $\pm$ 2.0 & 86.9 $\pm$ 0.4 & 65.8 $\pm$ 0.3 & 97.2 $\pm$ 0.3 & 23.2 $\pm$ 4.7 \\
Global Ens. & 41.4 $\pm$ 3.0 & 53.8 $\pm$ 0.5 & 57.0 $\pm$ 0.7 & 55.1 $\pm$ 1.4 & 89.6 $\pm$ 0.3 & 66.7 $\pm$ 1.9 & 93.1 $\pm$ 1.9 & 21.8 $\pm$ 1.0 \\
FedDES & \textbf{89.6} $\pm$ 1.8 & \textbf{77.7} $\pm$ 1.6 & \textbf{74.1} $\pm$ 1.4 & \textbf{57.3} $\pm$ 0.6 & \textbf{92.2} $\pm$ 0.2 & \textbf{73.1} $\pm$ 0.5 & \textbf{98.1} $\pm$ 0.7 & \textbf{25.1} $\pm$ 4.7 \\
\bottomrule
\end{tabular}}
\end{table}

\paragraph{Test performance.} Table~\ref{tab:main} reports mean test performance. FedDES achieves the highest score across all datasets, improving over the strongest baseline by 0.2 to 5.0 points. No single baseline is consistently the most competitive, with FML ranking second on CIFAR-10 and eICU, the Global Ensemble on BraTS and DomainNet, FedGH on scRNA-seq, and Local on PaySim. Relative to Local, FedDES gains the most on DomainNet (+11.1) and eICU (+6.9). Its smallest improvement (+0.7) comes on scRNA-seq, where Local already exceeds 97\% balanced accuracy and all pFL methods lie within 1.5 points of each other. No other baseline improves on Local on every dataset. FML comes closest, improving on all but PaySim. dFLMoE, the most closely related method, stays within 0.5 points of Local in every data setting.

The Global Ensemble's performance is highly sensitive to the type of heterogeneity. The cost of indiscriminate aggregation is most severe under the label skew of CIFAR-10, where many experts have seen few or none of the target client's classes and consistently assign them low probability. The combined probability mass of incompetent experts confidently predicting the wrong classes can therefore outweigh the mass competent experts place on the correct class. On the CIFAR-10 partitions, the Global Ensemble underperforms Local by 45.9, 19.0, and 11.5 points, with larger performance gaps corresponding to stronger label skew. Under the feature skew of BraTS and DomainNet, where every expert is exposed to the full label set during training, a larger fraction of the pool is likely to be competent for a given sample, and the remaining experts are less likely to be confidently incorrect. In these data settings, the Global Ensemble is the strongest baseline, outperforming Local by 2.6 and 8.9 points. A plausible contributing factor is that feature skew is a source of diversity in the expert pool, which, together with sufficient expert accuracy, drives ensemble gains. Label skew also diversifies the pool, but in the CIFAR-10 partitions, the skew is severe enough that competent experts are often outnumbered by those that are not, so errors are more likely to be shared than complementary.

\begin{table}[t]
\centering
\caption{Negative transfer rate (NTR, \%) and negative transfer burden (NTB, points), averaged over three seeds. Bold marks the lowest value within each data setting.}
\label{tab:ntr}
\resizebox{\textwidth}{!}{%
\begin{tabular}{l rr rr rr rr rr rr rr rr}
\toprule
& \multicolumn{6}{c}{CIFAR-10} & & & & & & & & & & \\
\cmidrule(lr){2-7}
& \multicolumn{2}{c}{Dir(0.1)} & \multicolumn{2}{c}{Dir(0.3)} & \multicolumn{2}{c}{Dir(0.5)} & \multicolumn{2}{c}{DomainNet} & \multicolumn{2}{c}{BraTS} & \multicolumn{2}{c}{eICU} & \multicolumn{2}{c}{scRNA-seq} & \multicolumn{2}{c}{PaySim} \\
\cmidrule(lr){2-3} \cmidrule(lr){4-5} \cmidrule(lr){6-7} \cmidrule(lr){8-9} \cmidrule(lr){10-11} \cmidrule(lr){12-13} \cmidrule(lr){14-15} \cmidrule(lr){16-17}
Method & NTR & NTB & NTR & NTB & NTR & NTB & NTR & NTB & NTR & NTB & NTR & NTB & NTR & NTB & NTR & NTB \\
\midrule
FedGH & 38.3 & 0.60 & 38.3 & 0.68 & 60.0 & 1.85 & 72.2 & 2.43 & 20.8 & 0.43 & 50.7 & 2.59 & 16.7 & 0.18 & 46.7 & 1.17 \\
FML & 15.0 & 0.21 & 8.3 & 0.08 & 11.7 & 0.26 & 19.4 & 0.22 & 20.8 & 0.32 & 32.0 & 1.37 & 16.7 & 0.11 & 60.0 & 1.87 \\
FedTGP & 16.7 & 0.32 & 13.3 & 0.34 & 23.3 & 0.61 & 22.2 & 0.75 & 54.2 & 1.71 & 66.7 & 7.33 & 16.7 & 0.19 & 80.0 & 4.49 \\
pFedMoE & 50.0 & 1.10 & 60.0 & 1.71 & 55.0 & 1.66 & 66.7 & 1.98 & 70.8 & 4.46 & 47.3 & 2.73 & 58.3 & 1.14 & 46.7 & 1.07 \\
dFLMoE & 23.3 & 0.30 & 30.0 & 0.55 & 41.7 & 0.90 & 38.9 & 1.25 & 37.5 & 0.50 & 45.3 & 2.06 & 33.3 & 0.33 & 60.0 & 1.20 \\
Global Ens. & 98.3 & 45.94 & 95.0 & 19.23 & 86.7 & 11.72 & 13.9 & 0.45 & 29.2 & 1.50 & 53.3 & 3.71 & 41.7 & 4.90 & 66.7 & 2.76 \\
FedDES & \textbf{3.3} & \textbf{0.02} & \textbf{3.3} & \textbf{0.03} & \textbf{0} & \textbf{0} & \textbf{0} & \textbf{0} & \textbf{8.3} & \textbf{0.06} & \textbf{15.3} & \textbf{0.61} & \textbf{8.3} & \textbf{0.07} & \textbf{13.3} & \textbf{0.27} \\
\bottomrule
\end{tabular}}
\end{table}
\begin{figure}[!t]
\centering
\includegraphics[width=\textwidth]{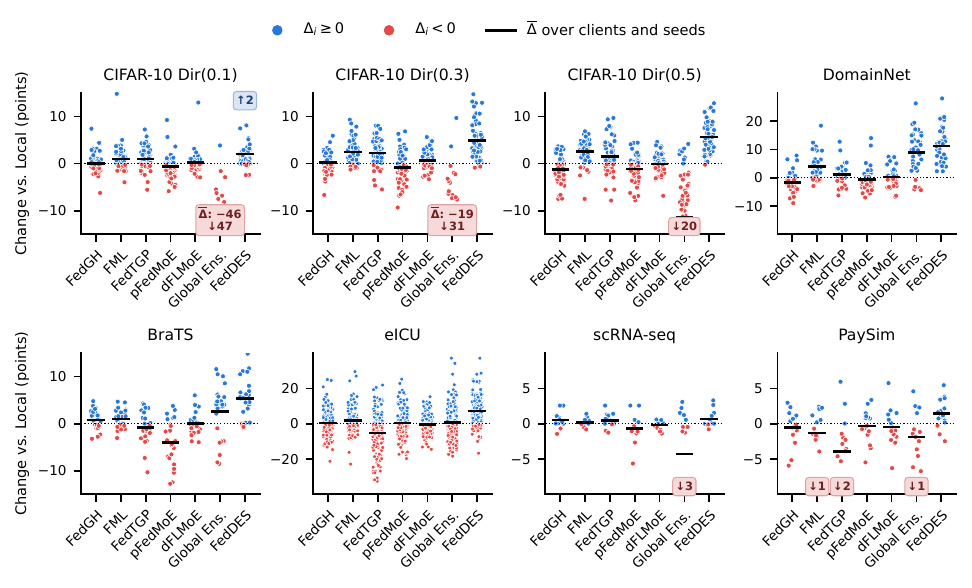}
\caption{Per-client difference in test performance from Local ($\Delta_i$). Boxes provide the number of clients with $\Delta_i$ beyond the axis range and the mean $\bar{\Delta}$ if it is also off-scale.}
\label{fig:client_gains_het}
\end{figure}

\paragraph{Client-level negative transfer.} Table~\ref{tab:ntr} reports NTR and NTB, and Figure~\ref{fig:client_gains_het} shows the underlying per-client gains. Read against Table~\ref{tab:main}, these results show that a baseline's mean improvement over Local is not a good indicator of whether the benefits of collaboration are broadly shared across clients. Even in the data settings where FML surpasses Local, its NTRs range from 8\% to 32\%. Its largest gain of 4.0 points on DomainNet comes with an NTR of 19\%, more than twice its NTR on CIFAR-10 Dir(0.3), where its margin is 2.5 points. FedGH, FedTGP, and the Global Ensemble each leave the majority of clients worse off in at least three data settings, and pFedMoE does so in five of eight, with NTRs ranging from 47\% to 71\%. On the CIFAR-10 partitions, the Global Ensemble's losses relative to Local extend to the large majority of clients, with NTRs of 98\%, 95\%, and 87\%. As with its mean performance gaps, the prevalence of negative transfer increases with the strength of label skew. By contrast, on DomainNet, its NTR of 14\% is lower than that of every method except FedDES. FedDES attains the lowest NTR and NTB on every dataset, with no more than 3.3\% of clients experiencing negative transfer on any CIFAR-10 partition and none on Dir(0.5) or DomainNet. Its highest NTR, 15\% on eICU, is still about half that of the next-best method.

\begin{table}[t]
\centering
\caption{Rates in the instance-level transfer framework. Each rate is the
share of a client's test samples with the given pattern of correct
(\checkmark) and incorrect ($\times$) predictions; a dash means the rate
does not involve the pool oracle. The personalized predictor is the
model a client uses at inference. The independent model $f_i$ is the
client's model trained without collaboration (the Local baseline), and
the pool $\mathcal{F}$ contains the independent models of all clients.
The pool oracle is correct whenever at least one model in $\mathcal{F}$ is.
FedDES and Global Ensemble combine these models directly; for the other
methods they serve only as a reference.}
\label{tab:framework}
\footnotesize
\begin{tabularx}{\textwidth}{@{}l ccc >{\raggedright\arraybackslash}X@{}}
\toprule
Rate & \shortstack{Independent\\model $f_i$} & \shortstack{Personalized\\predictor}
     & \shortstack{Pool\\oracle} & Reading \\
\midrule
Positive flip (PFR) & $\times$ & \checkmark & -- &
  Gain over independent training \\
\addlinespace
Negative flip (NFR) & \checkmark & $\times$ & -- &
  Unrealized gain; recoverable by deferring to $f_i$ \\
\addlinespace
Missed flip (MFR) & $\times$ & $\times$ & \checkmark &
  Unrealized gain; recoverable only by selecting another client's model \\
\addlinespace
Beyond pool (BPR) & $\times$ & \checkmark & $\times$ &
  Gain that no independent model could supply; attainable only through the personalized predictor (a subset of positive flips) \\
\midrule
\multicolumn{5}{@{}l@{}}{Gain over independent training: $\Delta_i = \mathrm{PFR}_i - \mathrm{NFR}_i$} \\
\multicolumn{5}{@{}l@{}}{Gap to Local-fallback oracle: $\mathrm{NFR}_i$ \qquad
  Gap to pool oracle: $\mathrm{NFR}_i + \mathrm{MFR}_i - \mathrm{BPR}_i$} \\
\bottomrule
\end{tabularx}
\end{table}

\begin{table}[!tb]
\centering
\caption{Instance-level transfer relative to Local (\%), averaged over clients and then over three seeds. The rates are defined in Table~\ref{tab:framework}. \textit{Pool oracle} selects a correct model from the clients' independently trained models whenever one exists, so its $\Delta$ is the most that such selection can gain over Local. A method's gap to the pool oracle is NFR $+$ MFR $-$ BPR. BPR is not shown; it is 0.0 for FedDES and the Global Ensemble and at most 1.1 points for the other methods.}
\label{tab:instance}
\footnotesize
\setlength{\tabcolsep}{1.5pt}
\resizebox{\textwidth}{!}{%
\begin{tabular}{@{}l *{5}{>{\hspace{3pt}}r r r r} @{}}
\toprule
& \multicolumn{4}{c}{\hspace{3pt}CIFAR-10 Dir(0.1)} & \multicolumn{4}{c}{\hspace{3pt}CIFAR-10 Dir(0.3)} & \multicolumn{4}{c}{\hspace{3pt}CIFAR-10 Dir(0.5)} & \multicolumn{4}{c}{\hspace{3pt}DomainNet} & \multicolumn{4}{c@{}}{\hspace{3pt}BraTS} \\
\cmidrule(l{4.5pt}r){2-5} \cmidrule(l{4.5pt}r){6-9} \cmidrule(l{4.5pt}r){10-13} \cmidrule(l{4.5pt}r){14-17} \cmidrule(l{4.5pt}r){18-21}
Method & \multicolumn{1}{>{\hspace{3pt}}c}{$\Delta$} & \multicolumn{1}{c}{PFR} & \multicolumn{1}{c}{NFR} & \multicolumn{1}{c}{MFR} & \multicolumn{1}{>{\hspace{3pt}}c}{$\Delta$} & \multicolumn{1}{c}{PFR} & \multicolumn{1}{c}{NFR} & \multicolumn{1}{c}{MFR} & \multicolumn{1}{>{\hspace{3pt}}c}{$\Delta$} & \multicolumn{1}{c}{PFR} & \multicolumn{1}{c}{NFR} & \multicolumn{1}{c}{MFR} & \multicolumn{1}{>{\hspace{3pt}}c}{$\Delta$} & \multicolumn{1}{c}{PFR} & \multicolumn{1}{c}{NFR} & \multicolumn{1}{c}{MFR} & \multicolumn{1}{>{\hspace{3pt}}c}{$\Delta$} & \multicolumn{1}{c}{PFR} & \multicolumn{1}{c}{NFR} & \multicolumn{1}{c@{}}{MFR} \\
\midrule
FedGH & $+0.0$ & 3.7 & 3.7 & 8.6 & $+0.2$ & 7.5 & 7.2 & 18.6 & $-1.3$ & 7.7 & 9.0 & 22.1 & $-1.6$ & 10.5 & 12.2 & 33.5 & $+0.8$ & 4.0 & 3.2 & 8.1 \\
FML & $+0.9$ & 3.6 & 2.6 & 8.8 & $+2.5$ & 7.4 & 5.0 & 18.6 & $+2.6$ & 8.3 & 5.7 & 21.4 & $+4.0$ & 12.6 & 8.6 & 31.1 & $+1.0$ & 3.9 & 3.0 & 8.2 \\
FedTGP & $+1.0$ & 4.4 & 3.4 & 7.9 & $+2.2$ & 8.3 & 6.0 & 17.8 & $+1.5$ & 8.8 & 7.3 & 21.0 & $+1.2$ & 11.4 & 10.2 & 32.5 & $-0.8$ & 4.3 & 5.0 & 7.9 \\
pFedMoE & $-0.6$ & 3.9 & 4.6 & 8.5 & $-0.8$ & 7.3 & 8.2 & 18.8 & $-1.2$ & 8.0 & 9.2 & 21.8 & $-0.7$ & 10.7 & 11.4 & 33.5 & $-4.0$ & 4.7 & 8.6 & 7.6 \\
dFLMoE & $+0.3$ & 2.9 & 2.6 & 9.4 & $+0.5$ & 6.0 & 5.5 & 20.1 & $-0.3$ & 6.4 & 6.7 & 23.4 & $+0.2$ & 9.9 & 9.7 & 34.1 & $-0.1$ & 3.3 & 3.4 & 8.9 \\
Global Ens. & $-45.9$ & 3.1 & 49.0 & 9.1 & $-19.0$ & 8.4 & 27.4 & 17.5 & $-11.4$ & 10.0 & 21.5 & 19.5 & $+8.9$ & 17.2 & 8.2 & 26.0 & $+2.6$ & 6.7 & 4.1 & 5.4 \\
FedDES & $+2.2$ & 4.0 & 1.7 & 8.3 & $+4.9$ & 9.1 & 4.2 & 16.8 & $+5.7$ & 10.5 & 4.8 & 19.0 & $+11.1$ & 17.6 & 6.4 & 25.6 & $+5.2$ & 7.7 & 2.5 & 4.5 \\
\addlinespace
\textit{Pool oracle} & $+12.2$ & 12.2 & 0.0 & 0.0 & $+25.9$ & 25.9 & 0.0 & 0.0 & $+29.5$ & 29.5 & 0.0 & 0.0 & $+43.2$ & 43.2 & 0.0 & 0.0 & $+12.1$ & 12.1 & 0.0 & 0.0 \\
\bottomrule
\end{tabular}}

\vspace{4pt}
\begin{minipage}{\textwidth}
\footnotesize
$\Delta$ is computed from unrounded accuracies and may differ by 0.1 from the difference of the rounded entries in Table~\ref{tab:main} and from PFR $-$ NFR.
\end{minipage}
\end{table}

\subsection{Instance-Level Transfer} Table~\ref{tab:framework} summarizes the instance-level transfer rates reported in Table~\ref{tab:instance}. 

\paragraph{Effect of selection.}
Comparing FedDES and the Global Ensemble, which differ only in how experts are weighted, isolates the effect of expert selection. Under the label skew of CIFAR-10, their PFRs are within one point of each other on every partition. However, the Global Ensemble's positive flips come at a much greater cost, with its NFR ranging from 21.5\% to 49.0\% across partitions, whereas FedDES's NFR stays below 5.0\%. Thus, FedDES's advantage in $\Delta$ over the Global Ensemble comes primarily from selection excluding experts that are poorly suited to a given sample, which under uniform weighting produce negative flips. The same pattern holds, to a lesser degree, under the feature skew of DomainNet and BraTS. The difference between their PFRs is comparable to that on CIFAR-10, within one point, but FedDES's NFR is lower by only 1.6 to 1.8 points.

\paragraph{Personalization mechanisms.}

Within a data setting, the average pairwise difference among the client-level pFL baselines (FedGH, FML, and FedTGP) is 0.7 points for PFR and 1.6 points for NFR. FML consistently has a lower NFR than FedGH and FedTGP, a pattern that may reflect how these methods integrate shared knowledge. FedGH replaces the client's prediction head with a global head before every round of local training, while FedTGP performs nearest-prototype classification using global class prototypes. Both largely confine personalization to the feature extractor, which must produce representations compatible with a shared classifier rather than one fitted to the local class distribution and within-class variation. FML, by contrast, retains a fully personalized model, incorporating shared knowledge solely through a distillation term in its training loss. In FedDES, which consistently has the lowest NFR, the independently trained local model remains fixed, serving as a source of purely local knowledge.

By weighting individual clients' prediction heads for each input, dFLMoE could in principle attain a lower MFR than methods that draw on consolidated knowledge. Instead, it has the highest MFR and lowest PFR of any method in all five data settings. A likely reason is that the received heads operate on the target client's representations through locally trained transformations. These transformations cannot recover distinctions not encoded by the local feature extractor and have limited training data for underrepresented classes or domains, where source knowledge may be most useful. Consequently, the received heads may serve more as alternative classifiers of local representations than as carriers of source knowledge. Consistent with this interpretation, even the Global Ensemble, which performs no selection, consistently has a higher PFR and lower MFR than dFLMoE.

pFedMoE adaptively mixes the outputs of the local and shared feature extractors for each input and would therefore be expected to attain a lower NFR than the client-level baselines. Yet its NFR is the highest among the pFL baselines in four of the five data settings. Because the local and shared extractors are trained jointly and the prediction head is fitted to their mixed representations, relying exclusively on the local extractor does not recover the independently trained model's prediction. Together, these results suggest that the benefits of instance-level adaptation depend on the extent to which local knowledge is preserved, while those of selective integration depend on how much of each source client's knowledge is transferred.

\paragraph{Unrealized gain.} The personalized predictors of the pFL baselines are trained jointly with other clients, at least in part, and may therefore generalize in ways that no independently trained model or combination of them can. If this ability accounted for much of their gain, it would weigh against FedDES's approach of selecting among independently trained models. However, the baselines' BPRs are at most 1.1 points, small relative to their PFRs of 2.9 to 12.6. Nearly all their positive flips therefore occur on samples that some independently trained model already classifies correctly, leaving NFR and MFR to account for almost the entire gap to the pool oracle.

For FedGH, FML, and FedTGP, which personalize at the client level by drawing on consolidated knowledge, NFR is the portion of this gap recoverable through instance-level personalization and MFR the portion recoverable only through selective integration of individual source clients' knowledge. MFR exceeds NFR in every data setting by a factor of 1.6 to 3.8, so most of the gain left unrealized would require selective knowledge integration. Consistent with this, FedDES has the lowest MFR in every data setting except CIFAR-10 Dir(0.1), where FedTGP's is 0.4 points lower.

\subsection{Additional Analyses and Ablations}

\begin{wraptable}{r}{0.5\textwidth}
\vspace{-18pt}
\centering
\caption{FedDES routing sparsity over a pool of 20 experts on the CIFAR-10 partitions, averaged over each client's test samples and then over clients (mean $\pm$ SD over three seeds). Experts selected: the number of experts with $\alpha_j(\mathbf{x}) > 0$. Effective ensemble size: $1/\sum_j \alpha_j(\mathbf{x})^2$. Local expert's weight: the routing weight on the client's own expert.}
\label{tab:ensemble}
\footnotesize\setlength{\tabcolsep}{3pt}
\begin{tabular}{l ccc}
\toprule
& Dir(0.1) & Dir(0.3) & Dir(0.5) \\
\midrule
Experts selected & 8.1 $\pm$ 0.2 & 14.5 $\pm$ 1.4 & 17.2 $\pm$ 1.0 \\
Effective ensemble size & 5.0 $\pm$ 0.6 & 8.9 $\pm$ 0.8 & 11.2 $\pm$ 1.0 \\
Local expert's weight & 0.27 $\pm$ 0.02 & 0.16 $\pm$ 0.02 & 0.13 $\pm$ 0.02 \\
\bottomrule
\end{tabular}
\end{wraptable}
\paragraph{Routing sparsity.} Table~\ref{tab:ensemble} summarizes how FedDES distributes weight over the pool of 20 experts on the CIFAR-10 partitions. Routing becomes sparser as heterogeneity increases, with FedDES assigning nonzero weight to an average of 17.2 experts per sample on Dir(0.5),  14.5 on Dir(0.3), and 8.1 on Dir(0.1). The average effective ensemble size, which accounts for weight concentration among selected experts, is smaller and in a narrower range (5.0--11.2), but follows the same trend. This is consistent with the behavior of the Global Ensemble, whose performance degrades with stronger label skew as fewer experts are competent on a given client's samples. The weight on the client's own expert rises from 0.13 on Dir(0.5) to 0.27 on Dir(0.1), compared with 0.05 under uniform weighting, yet over 70\% of the weight still goes to other clients' experts, so FedDES does not reduce to local training.

\begin{figure}[t]
\centering
\includegraphics[width=\textwidth]{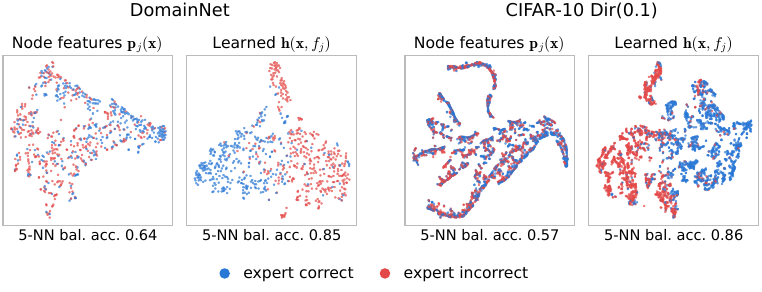}
\caption{t-SNE projections of sample--expert pairs on DomainNet and CIFAR-10 Dir(0.1), colored by whether the expert classifies the sample correctly, using the node features and the router's learned representations. Values give the balanced accuracy of a 5-nearest-neighbor classifier that predicts expert correctness from each representation.}
\label{fig:tsne}
\end{figure}

\paragraph{Learned pair representations.} Figure~\ref{fig:tsne} shows t-SNE projections of sample--expert pairs on DomainNet and CIFAR-10 Dir(0.1) before and after message passing. For each dataset, the projections come from the client with the most test samples on which some experts are correct and others are not, and the predicted class whose pairs are most evenly split between correct and incorrect experts (Appendix~\ref{app:tsne}). A 5-nearest-neighbor classifier predicts expert correctness from the initial node features with a balanced accuracy of only 0.64 on DomainNet and 0.57 on CIFAR-10 Dir(0.1). Using the learned representations, its balanced accuracy rises to 0.85 and 0.86. The router therefore learns a representation space in which a sample--expert pair's position is more informative of expert competence.

\begin{wraptable}{r}{0.4\textwidth}
\vspace{-18pt}
\centering
\caption{Training cost on CIFAR-10 Dir(0.3), summed over clients up to each run's early stop and averaged over three seeds. FedDES includes pool and router training; w/o CV fits excludes the cross-validation fits.}
\label{tab:resources}
\footnotesize
\begin{tabular}{l rr}
\toprule
& Communication & FLOPs \\
Method & (GiB) & (PFLOPs) \\
\midrule
FedGH & 0.10 & 16.8 \\
FML & 40.2 & 29.6 \\
FedTGP & 0.18 & 30.9 \\
pFedMoE & 27.1 & 15.5 \\
dFLMoE & 1.44 & 14.1 \\
FedDES & 6.97 & 44.0 \\
\quad w/o CV fits & 6.97 & 10.2 \\
\bottomrule
\end{tabular}
\end{wraptable}
\paragraph{Communication and computation costs.} Table~\ref{tab:resources} reports training costs on CIFAR-10 Dir(0.3). The communication cost of FedDES's one-time model exchange is 4 to 6 times lower than that of methods that exchange model parameters every round (FML and pFedMoE). Methods that share only classifier heads or prototypes (FedGH, FedTGP, and dFLMoE) have the lowest communication costs, ranging from 0.1 to 1.4~GiB. FedDES has the highest training cost (44.0~PFLOPs), though about three quarters of it comes from cross-validation fits. The cross-validation fits are not inherent to the FedDES algorithm, but rather an experimental design choice that lets each client use all of its training data to train both its expert and its router. Without these fits, FedDES has the lowest training cost (10.2~PFLOPs), of which router training accounts for only about 6\%.

\wrapfill

\paragraph{Ablations.}
To isolate the contribution of each component of FedDES,
Table~\ref{tab:ablation} compares it with three variants on
the CIFAR-10 partitions, all using the same expert pool.
\emph{FedDES-static} removes per-sample routing, instead
fitting a single weight vector over the experts for each
client by minimizing the log loss of the weighted soft
vote on the client's training samples. The \emph{MLP router}
retains per-sample routing but removes message passing by
replacing the GNN with an MLP of the same depth and hidden
dimension. The \emph{heterogeneous graph} variant retains
message passing but uses separate nodes for samples and
experts rather than nodes representing sample--expert pairs
(Section~\ref{sec:graph}). Each expert's predictions and
correctness status are encoded as edge features, and each
sample--expert pair is scored from the concatenated
embeddings of its sample and expert nodes.
On every partition, accuracy rises in order from FedDES-static to the MLP router, then the heterogeneous graph, then FedDES. Per-sample routing alone adds 0.6 to 1.5 points over FedDES-static, and message passing adds a further 1.2 to 2.3 points over the MLP router. Most of this further gain depends on treating sample--expert pairs as nodes, since message passing over the heterogeneous graph adds only 0.2 to 0.7 points. FedDES's negative transfer is also as low as any variant's. Its NTR and NTB are each the lowest or tied for lowest on every partition.

Table~\ref{tab:ablation_instance} shows how each component changes the composition of FedDES's gain over local training. On every partition, FedDES attains the highest PFR and leaves the least gain unrealized relative to the pool oracle, with the lowest sum of NFR and MFR. Compared with the MLP router, message passing over sample--expert pairs raises PFR by 1.6 to 2.3 points while increasing NFR by at most 0.5. Relative to the heterogeneous graph, it raises PFR by 0.8 to 1.0 points without increasing NFR. For every variant, MFR accounts for most of the remaining gap. Further gains would therefore come mainly from more selective integration of source experts' knowledge.

\begin{table}[!ht]
\centering
\caption{Ablation on the CIFAR-10 partitions. Acc.: test accuracy (mean $\pm$ SD over three seeds). NTR (\%) and NTB (points) are defined as in Table~\ref{tab:ntr}. Bold marks the best value in each column.}
\label{tab:ablation}
\footnotesize\setlength{\tabcolsep}{4pt}
\begin{tabular}{l ccc ccc ccc}
\toprule
& \multicolumn{3}{c}{Dir(0.1)} & \multicolumn{3}{c}{Dir(0.3)} & \multicolumn{3}{c}{Dir(0.5)} \\
\cmidrule(lr){2-4}\cmidrule(lr){5-7}\cmidrule(lr){8-10}
Variant & Acc. & NTR & NTB & Acc. & NTR & NTB & Acc. & NTR & NTB \\
\midrule
FedDES & \textbf{89.56 $\pm$ 1.79} & \textbf{3.3} & \textbf{0.02} & \textbf{77.69 $\pm$ 1.60} & \textbf{3.3} & \textbf{0.03} & \textbf{74.14 $\pm$ 1.45} & \textbf{0} & \textbf{0} \\
Heterogeneous graph & 88.53 $\pm$ 2.16 & 13.3 & 0.15 & 76.33 $\pm$ 1.67 & 10.0 & 0.16 & 72.57 $\pm$ 1.20 & 5.0 & 0.06 \\
MLP router & 88.36 $\pm$ 2.27 & 8.3 & 0.15 & 75.92 $\pm$ 1.52 & 11.7 & 0.14 & 71.84 $\pm$ 1.71 & 5.0 & 0.06 \\
FedDES-static & 87.72 $\pm$ 2.31 & 5.0 & 0.03 & 74.45 $\pm$ 2.31 & \textbf{3.3} & \textbf{0.03} & 71.07 $\pm$ 1.27 & \textbf{0} & \textbf{0} \\
\bottomrule
\end{tabular}
\end{table}

\begin{table}[!ht]
\centering
\caption{Instance-level transfer of the ablation variants relative to Local (\%) on the CIFAR-10 partitions, averaged over clients and then over three seeds. The rates are defined in Table~\ref{tab:framework}.}
\label{tab:ablation_instance}
\footnotesize\setlength{\tabcolsep}{3pt}
\begin{tabular}{l rrrr rrrr rrrr}
\toprule
& \multicolumn{4}{c}{Dir(0.1)} & \multicolumn{4}{c}{Dir(0.3)} & \multicolumn{4}{c}{Dir(0.5)} \\
\cmidrule(lr){2-5}\cmidrule(lr){6-9}\cmidrule(lr){10-13}
Variant & $\Delta$ & PFR & NFR & MFR & $\Delta$ & PFR & NFR & MFR & $\Delta$ & PFR & NFR & MFR \\
\midrule
FedDES & $+2.2$ & 4.0 & 1.7 & 8.3 & $+4.9$ & 9.1 & 4.2 & 16.8 & $+5.7$ & 10.5 & 4.8 & 19.0 \\
Heterogeneous graph & $+1.2$ & 3.2 & 2.0 & 9.0 & $+3.5$ & 8.2 & 4.7 & 17.7 & $+4.1$ & 9.5 & 5.4 & 20.0 \\
MLP router & $+1.0$ & 2.4 & 1.4 & 9.8 & $+3.1$ & 6.8 & 3.7 & 19.1 & $+3.4$ & 8.2 & 4.8 & 21.4 \\
FedDES-static & $+0.4$ & 0.9 & 0.5 & 11.3 & $+1.6$ & 2.8 & 1.2 & 23.1 & $+2.6$ & 4.3 & 1.7 & 25.3 \\
\bottomrule
\end{tabular}
\end{table}

\section{Practical Considerations}

\paragraph{Privacy implications.}
FedDES's model exchange protocol shares many of the same privacy considerations as other methods within the model-sharing FL paradigm. In \citeauthor{shao2023survey}'s \citeyearpar{shao2023survey} comparison of FL methods by the type of information shared, model sharing is characterized as offering the greatest model utility but also having greater exposure to white-box privacy attacks. Such attacks include membership inference, property inference, and training data reconstruction \citep{shokri2017membership,ganju2018property,haim2022reconstructing}. The relevance of specific classes of attacks varies with the type of model information exchanged. Gradient-sharing methods are susceptible to gradient inversion attacks, which reconstruct training samples from shared gradients \citep{zhu2019deep,geiping2020inverting,zhao2020idlg}, whereas methods that iteratively share intermediate model states are exposed to attacks that exploit parameter updates across training rounds to infer sample membership or other sensitive attributes \citep{melis2019exploiting}. While neither of these applies to FedDES's model exchange protocol, training data reconstruction from a trained model's parameters remains a risk \citep{haim2022reconstructing,buzaglo2023reconstructing}. Several established privacy-preserving mechanisms are compatible
with this setting, including training with differential privacy
\citep{abadi2016deep} to limit information leakage from shared
models and the use of trusted execution environments
\citep{intel2016sgx} to enable model evaluation without direct
access to model parameters.

\paragraph{Applicability across federated settings.}
Although FedDES is not restricted to cross-silo FL, several
characteristics of its design make it particularly well suited
to this setting. Because FedDES draws on knowledge encoded in independently trained models rather than learning shared representations through collaborative training, its ability to improve on local training depends on the availability of suitable experts across the target client's input space. In cross-silo FL, clients often hold sufficient data to train models that are effective in at least some regions of the input space. Given differences in clients' data distributions, these regions may differ across models, providing a basis for expert specialization. In this way, the statistical heterogeneity that poses a learning challenge for methods that aggregate cross-client knowledge can actually broaden the coverage of the expert pool in FedDES. Our instance-level transfer analysis shows that relatively few of the gains achieved by collaboratively trained predictors fall beyond the reach of the independently trained models in our evaluated settings. In cross-device FL, where individual clients typically hold much smaller datasets, collaborative training may offer greater opportunities to learn knowledge that is not captured by any independently trained model.

Cross-silo FL is also favorable from a computational standpoint, as it typically involves fewer clients with greater computational resources. Although FedDES requires only a one-time model exchange, each client maintains the full expert pool and evaluates all experts at inference, regardless of routing sparsity. Moreover, many of the applications that motivate cross-silo FL, such as healthcare, involve high-stakes decisions for which predictive performance is often the primary consideration when choosing among computationally feasible methods.

\section{Conclusion}
This paper introduces FedDES, an instance-level personalized federated learning method in which clients dynamically select and weight a relevant subset of independently trained client models (i.e., experts) for each input. Expert selection is performed by a GNN operating on a graph over sample--expert pairs, in which sample similarity provides the relational structure for learning representations predictive of expert competence. Across eight data settings with both simulated and naturally occurring heterogeneity, FedDES outperforms existing model-heterogeneous pFL methods while more effectively mitigating negative transfer. We also introduce an instance-level transfer framework that provides insight into how different personalization strategies perform relative to independent training and quantifies the potential gains from instance-level personalization and selective knowledge integration. Our findings suggest that the benefits of finer-grained personalization depend not only on how knowledge is integrated but also on how that knowledge is represented.

\FloatBarrier
\wrapfill

\bibliographystyle{tmlr}
\bibliography{bibliography}

\clearpage
\appendix
\section*{Appendix}

\section{Router Architecture}
\label{app:arch}

Here we detail the router architecture summarized in
Section~\ref{sec:router}, along with its training procedure. At layer $t$, $\mathbf{h}(\cdot)$ denotes
$\mathbf{h}^{(t-1)}(\cdot)$. All weight matrices are learned separately for each layer, and bias terms are omitted for simplicity. We use single-head attention.

\paragraph{Competence message.}
We compute the competence message using the graph
transformer convolution of \citet{shi2021masked}. The edge weight is embedded as $\mathbf{e}(\mathbf{x}', \mathbf{x}) = w(\mathbf{x}', \mathbf{x})\,\mathbf{w}_E$, where $\mathbf{w}_E \in \mathbb{R}^d$ is a learned vector. The edge embedding is then added to the keys and values: 
\begin{gather*}
\beta_{\mathbf{x}'} = \operatorname*{softmax}_{\mathbf{x}' \in \mathcal{R}(\mathbf{x})}
\left( \frac{\big(\mathbf{W}^{\mathrm{c}}_Q\, \mathbf{h}(\mathbf{x}, f_j)\big)^{\top}
\big(\mathbf{W}^{\mathrm{c}}_K\, \mathbf{h}(\mathbf{x}', f_j) + \mathbf{e}(\mathbf{x}', \mathbf{x})\big)}{\sqrt{d}} \right), \\
\mathbf{m}^{\mathrm{comp}}(\mathbf{x}, f_j) = \sum_{\mathbf{x}' \in \mathcal{R}(\mathbf{x})} \beta_{\mathbf{x}'}
\big(\mathbf{W}^{\mathrm{c}}_V\, \mathbf{h}(\mathbf{x}', f_j) + \mathbf{e}(\mathbf{x}', \mathbf{x})\big).
\end{gather*}

\paragraph{Layer update.}
The message is combined with the node representation
through a residual connection followed by layer normalization.
The result is passed through a feed-forward block with its
own residual connection and layer normalization:
\begin{gather*}
\tilde{\mathbf{h}} = \mathrm{LN}_1\big(\mathbf{h} + \mathbf{m}^{\mathrm{comp}}\big), \\
\mathbf{h}^{(t)} = \mathrm{LN}_2\big(\tilde{\mathbf{h}} + \mathbf{W}_2\,\mathrm{ReLU}(\mathbf{W}_1 \tilde{\mathbf{h}})\big),
\end{gather*}
where all terms are evaluated at node $(\mathbf{x}, f_j)$,
$\mathrm{LN}$ denotes layer normalization, and the
feed-forward block has a hidden dimension of $2d$.
Dropout is applied to $\mathbf{h}^{(t)}$ between layers.

\paragraph{Training.}
Algorithm~\ref{alg:router} outlines the full-batch
training procedure for the router. On eICU, PaySim,
and scRNA-seq, $\omega(\mathbf{x})$ denotes the
inverse-frequency class weight; otherwise,
$\omega(\mathbf{x}) = 1$. The learning rate is halved when validation performance has not improved for 25 epochs.

\begin{algorithm}[!ht]
\caption{Router training at a single client}
\label{alg:router}
\begin{algorithmic}[1]
\Require graph $\mathcal{G}$ over $\mathcal{D}^{\mathrm{tr}}$ and $\mathcal{D}^{\mathrm{val}}$, masking probability $\rho$, maximum epochs $E$, early-stopping patience $P$
\State initialize $\phi$; $v^{*} \gets -\infty$
\For{$e = 1, \dots, E$}
    \State $\mathcal{M} \gets$ random subset of $\mathcal{D}^{\mathrm{tr}}$ \Comment{each sample independently with probability $\rho$}
    \State $\mathcal{G}_{\mathcal{M}} \gets \textsc{Mask}(\mathcal{G}, \mathcal{M})$ \Comment{Section~\ref{sec:training}}
    \State $\mathcal{M}^{\circ} \gets \{\mathbf{x} \in \mathcal{M} : 0 < |\mathcal{J}(\mathbf{x})| < N\}$
    \State $\ell \gets \frac{1}{|\mathcal{M}^{\circ}|} \sum_{\mathbf{x} \in \mathcal{M}^{\circ}} \omega(\mathbf{x})\, L(\mathbf{x})$ \Comment{forward pass on $\mathcal{G}_{\mathcal{M}}$}
    \State $\phi \gets \textsc{Adam}(\phi, \nabla_{\phi}\,\ell)$
    \State $v \gets$ validation metric of $\mathbf{g}$ \Comment{forward pass on $\mathcal{G}$}
    \If{$v > v^{*}$}
        \State $v^{*} \gets v$, \ $\phi^{*} \gets \phi$
    \ElsIf{$v^{*}$ has not improved for $P$ epochs}
        \State \textbf{break}
    \EndIf
\EndFor
\State \Return $\phi^{*}$
\end{algorithmic}
\end{algorithm}

\clearpage
\section{Additional Results}
\label{app:results}

\subsection{Learned Pair Representations}
\label{app:tsne}

For each dataset in Figure~\ref{fig:tsne}, we select the client with the largest number of test samples on which some, but not all, experts are correct. To avoid clustering by predicted class, we restrict the visualization to sample--expert pairs with the same predicted class, selecting the class with the most balanced distribution of correct and incorrect predictions. This yields 700 pairs on DomainNet (335 correct predictions) and 2,878 pairs on CIFAR-10 (1,519 correct predictions).

The 10-dimensional probability vectors $\mathbf{p}_j(\mathbf{x})$ and the 128-dimensional representations $\mathbf{h}^{(T)}(\mathbf{x}, f_j)$ are projected separately using t-SNE with a perplexity of 30 and PCA initialization. We evaluate the 5-nearest-neighbor classifier in the original representation spaces using 5-fold cross-validation, grouped by test sample, and report the mean balanced accuracy.

\section{Dataset Preprocessing}
\label{app:datasets}

\paragraph{CIFAR-10.} Images are scaled to $[0,1]$ and normalized using the channel-wise means and standard deviations. We impose a minimum client dataset size of 250 samples during partitioning.

\paragraph{BraTS.} Each of the four MRI modalities (T1, T1Gd, T2, and FLAIR) is clipped at its 0.5th and 99.5th percentiles and z-score normalized over brain voxels, with the background set to zero. We retain every second axial slice in which at least
1\% of the pixels correspond to brain tissue and resize the retained slices to $128 \times 128$ using area interpolation. A slice is labeled positive if it contains any tumor voxel. The resulting dataset comprises 1,107 subjects. Within each institution, test subjects are selected to preserve the distribution of the proportion of tumor-containing slices per subject.

\paragraph{DomainNet.} Images are center-cropped to a square, resized to $128 \times 128$ using bilinear interpolation, and scaled to $[0,1]$. We impose a minimum client dataset size of 500 samples during partitioning.

\paragraph{eICU.} We include adult ICU stays with at least one time-series measurement, a known in-hospital outcome, and a complete 48-hour observation window. The time series comprise 43 clinical variables, each paired with a decaying
missingness indicator. Variables are retained if measured in at least 25\% of patients in at least 70\% of hospitals. Values are scaled using the 5th and 95th percentiles to map this range to $[-1,1]$, clipped at $\pm 4$, and forward-filled, with remaining missing values set to zero. The static features include 47 admission
features and 271--276 diagnosis indicators derived from active diagnoses, admission diagnoses, and past medical history recorded in the first 5 hours. Diagnosis indicators are retained if present in at least 1\% of stays. Hospital-level variables and APACHE variables available only after 24 hours are excluded. All feature selection and preprocessing statistics are based on the training data only. 

\paragraph{scRNA-seq.} We use the datasets as released with the benchmark of \citet{abdelaal2019comparison}, harmonize cell-type names across studies, and restrict genes to those shared by all four studies. Cells flagged by Segerstolpe's quality control are removed. A cell type is excluded from a study if it has fewer than 50 cells in that study, and excluded altogether if it has fewer than 10 cells across studies. Expression values are normalized to counts per million and log-transformed. The 2,000 most variable genes across the pooled cells are retained, and each gene is min--max scaled separately for each study. Cell-type filtering, gene selection, and scaling statistics are based on the training data only.

\paragraph{PaySim.} The input features are the time step and transaction amount, standardized using statistics from the source training data, and a one-hot encoding
of the five transaction types. We exclude the four account-balance features because the simulator cancels detected fraudulent transactions, introducing label leakage through the resulting balances. Account identifiers and the simulator's fraud flag are also excluded.

\section{Model Architectures}
\label{app:models}

Each client model consists of a feature extractor followed
by a linear classifier. All image models are trained from
scratch.

\paragraph{Image models.} The CNN is adapted from \citet{mcmahan2017communication}
and consists of two $5 \times 5$ convolutional layers
with 32 and 64 filters, each followed by ReLU and
$2 \times 2$ max pooling. The output of the
second block is passed through an adaptive average pooling
layer with an output size of $5 \times 5$ and a fully
connected layer with 512 units. ResNet-18~\citep{he2016deep} and MobileNetV2~\citep{sandler2018mobilenetv2} produce 512 and 1,280 features, respectively. On CIFAR-10 ($3 \times 32 \times 32$ inputs), both use the small-image stem common at this resolution. ResNet-18 uses a $3 \times 3$ stride-1 convolution in place of the initial $7 \times 7$ stride-2 convolution and omits the initial max pooling layer. MobileNetV2 uses stride 1 in the initial convolution
and second inverted residual stage, and replaces
ReLU6 with ReLU. For BraTS ($4 \times 128 \times 128$) and DomainNet ($3 \times 128 \times 128$), we retain the standard stems and use a dropout rate of 0.2 before the MobileNetV2 classifier.

\paragraph{eICU models.}
Each model receives the hourly time series and static features described in Appendix~\ref{app:datasets}. The GRU model~\citep{cho2014learning} encodes the time series with a single-layer GRU of hidden dimension 64 and projects the static features to 64 dimensions. The concatenated representations are projected to 128 dimensions. TPC-light is a reduced variant of the temporal pointwise convolution (TPC) network of \citet{rocheteau2021temporal}, with three TPC layers, a temporal kernel size of 4, four kernels per variable, dilations of 1, 3, and 6, and a pointwise dimension of 8. Dropout rates are 0.05 (temporal) and 0.45 (pointwise and final). The output is averaged over time, combined with the static features, and projected to 64 dimensions. The feed-forward attention model~\citep{raffel2015feedforward} projects each time step to 64 dimensions and applies attention pooling over time. The result is combined with a 64-dimensional encoding of the static features and projected to 64 dimensions, with a dropout rate of 0.3 on the output.

\paragraph{Tabular models.}
For scRNA-seq (2,000 input genes), the MLP consists of input batch normalization and a fully connected layer with 64 units, followed by batch normalization and ReLU. The gated MLP uses input batch normalization, a fully connected layer with 64 units, a gated linear unit (GLU)~\citep{dauphin2017language}, and a projection to 64 dimensions. For PaySim (7 input features), the MLP has two fully connected layers with 128 and 64 units, each followed by layer normalization, ReLU, and dropout with a rate of 0.1. DCN-V2~\citep{wang2021dcnv2} adds two cross layers before the same MLP. TabNet~\citep{arik2021tabnet} uses decision and attention dimensions of 8, three decision steps, a relaxation factor of 1.3, two shared and two step-specific GLU layers, and a virtual batch size of 128. We omit the sparsity loss and set the output dimension to 64.

\paragraph{Adaptations for FedGH, FedTGP, and pFedMoE.} The common feature dimension $m$ is set to 512 for the image datasets, 128 for eICU, and 64 for scRNA-seq and PaySim. For FedGH and pFedMoE, a learned linear projection maps each feature extractor's output to $m$ dimensions. In pFedMoE, the shared expert has its own projection, and the gating network has a hidden layer of 32 units with layer normalization and sigmoid activation, followed by a two-way softmax. For FedTGP, adaptive average pooling reduces the 1,280-dimensional MobileNetV2 features to 512 dimensions and expands the 64-dimensional outputs of TPC-light and the attention model to 128 dimensions by repeating each feature twice. Models whose output dimension already matches $m$ are unchanged.

\section{Hyperparameters}
\label{app:hyperparams}

Tables~\ref{tab:hp-feddes} and~\ref{tab:hp-baselines}
summarize the hyperparameters used for FedDES and the baselines.

\begin{table}[!htb]
\centering
\caption{FedDES router hyperparameters. Values in braces
denote the hyperparameter search space.}
\label{tab:hp-feddes}
\small
\begin{tabular}{ll}
\toprule
Hyperparameter & Value \\
\midrule
Message-passing layers $T$ & 2 \\
Neighbors $k$ & $\{7, 15, 25\}$ \\
Hidden dimension $d$ & $\{128, 256\}$ \\
Masking probability $\rho$ & 0.2 \\
Entmax order & 1.5 \\
Optimizer & Adam \\
Learning rate & $10^{-3}$ \\
Weight decay & $10^{-4}$ \\
Maximum epochs & 500 \\
Early-stopping patience & 50 \\
\bottomrule
\end{tabular}

\vspace{4pt}
{\footnotesize On PaySim, $d = 256$ is evaluated only
with $k = 7$ due to GPU memory constraints.}
\end{table}

\begin{table}[!htb]
\centering
\caption{Method-specific hyperparameters of the baselines.
Shared training settings are given in
Section~\ref{sec:experiments}.}
\label{tab:hp-baselines}
\footnotesize
\setlength{\tabcolsep}{5pt}
\begin{tabular}{@{}llll@{}}
\toprule
Method & Hyperparameter & Value & Source \\
\midrule
FedGH
    & Global head learning rate & 0.01 & Original \\
    & Server head updates & One pass per round & Original \\
\midrule
FedTGP
    & Prototype loss weight $\lambda$ & $\{0.01, 0.1, 1\}$ & Tuned \\
    & Server learning rate & 0.01 & Original \\
    & Server epochs $S$ & 100 & Original \\
    & Margin cap $\tau$ & 100 & Original \\
\midrule
FML
    & Loss weights $\alpha, \beta$ & 0.5, 0.5 & Ours \\
\midrule
pFedMoE
    & Gate hidden dimension & 32 & Ours \\
    & Gate learning rate & 0.01 & Ours \\
    & Shared extractor learning rate & 0.01 & Original \\
\midrule
dFLMoE
    & Feature-transform and router learning rate
    & $\{10^{-4}, 10^{-3}, 10^{-2}\}$ & Tuned \\
    & Loss weights $\lambda_{\mathrm{loc}}, \lambda_{\mathrm{MoE}}$
    & 0.5, 0.5 & Original \\
    & Common feature dimension & 128 & Ours \\
    & Router-stage epochs per round & 1 & Ours \\
\bottomrule
\end{tabular}

\vspace{4pt}
\begin{minipage}{\textwidth}
\footnotesize
Original: values reported in the original papers.
Tuned: values selected from the indicated search space.
Ours: values not specified in the original papers.
\end{minipage}
\end{table}

\paragraph{Unspecified hyperparameters.}
For FML, we set $\alpha = \beta = 0.5$, weighting the
cross-entropy and distillation terms equally, as in deep
mutual learning~\citep{zhang2018deep}, on which FML builds.
For pFedMoE, we use a gate hidden dimension of 32 and
the same learning rate as the shared feature extractor.
For dFLMoE, we set the common feature dimension to 128
and use one epoch per round for the second training
stage, matching the first stage.

\paragraph{Learning-rate adjustments.}
For the few method--dataset combinations where training
diverged, we reduced the learning rate across all three
seeds. For FedGH, we reduce the learning rate from 0.01
to 0.003 on CIFAR-10 Dir(0.1) and DomainNet.
For FedTGP on PaySim, we use 0.001 after observing
divergence at 0.01 and 0.003.
For dFLMoE on CIFAR-10 Dir(0.3), the selected
feature-transform and router learning rate of 0.01
caused divergence on one seed, so we use the next-best
value of 0.001.

\end{document}